\pdfoutput=1

\documentclass{article}

\PassOptionsToPackage{numbers, compress}{natbib}

\usepackage[preprint]{neurips_2024}

\usepackage[utf8]{inputenc} 
\usepackage[T1]{fontenc}    
\usepackage[dvipsnames]{xcolor}
\usepackage{hyperref}       
\hypersetup{
    colorlinks=true,
	citecolor=MidnightBlue,
	linkcolor=OrangeRed,
	urlcolor=OrangeRed
}
\usepackage{url}            
\usepackage{booktabs}       
\usepackage{amsfonts}       
\usepackage{nicefrac}       
\usepackage{microtype}      
\usepackage{xcolor}         

\usepackage{todonotes}
\usepackage{amsthm}
\usepackage{amsmath}
\usepackage[ruled,vlined]{algorithm2e}
\usepackage[capitalise,nameinlink]{cleveref}
\usepackage{amssymb}
\usepackage{makecell}
\usepackage{multirow}
\usepackage{footnote}
\usepackage{subcaption}
\usepackage{graphicx}
\usepackage{wrapfig}

\usepackage{amsmath,amsfonts,bm}

\def\eqref#1{equation~\ref{#1}}

\def\1{\bm{1}}

\def\vb{{\bm{b}}}

\def\vn{{\bm{n}}}

\def\vv{{\bm{v}}}

\def\vx{{\bm{x}}}

\def\evx{{x}}

\def\mI{{\bm{I}}}
\def\mJ{{\bm{J}}}

\def\mM{{\bm{M}}}

\def\mT{{\bm{T}}}
\def\mU{{\bm{U}}}

\def\mW{{\bm{W}}}

\DeclareMathAlphabet{\mathsfit}{\encodingdefault}{\sfdefault}{m}{sl}
\SetMathAlphabet{\mathsfit}{bold}{\encodingdefault}{\sfdefault}{bx}{n}

\def\sS{{\mathbb{S}}}

\newcommand{\R}{\mathbb{R}}

\newtheorem{definition}{Definition}

\newtheorem{proposition}{Proposition}
\newtheorem*{remark}{Remark}

\title{Fast Shapley Value Estimation: A Unified Approach}

\newcommand*\samethanks[1][\value{footnote}]{\footnotemark[#1]}
\author{%
  Borui Zhang\thanks{Equal contribution.}\ , Baotong Tian\samethanks\ , Wenzhao Zheng\ , Jie Zhou\ , Jiwen Lu\thanks{Corresponding author.}\\
  {Department of Automation, Tsinghua University, China}\\
  \texttt{\{zhang-br21, tbt20, zhengwz18\}@mails.tsinghua.edu.cn; \{jzhou, lujiwen\}@tsinghua.edu.cn} \\
}

\begin{document}

\maketitle

\begin{abstract}

Shapley values have emerged as a widely accepted and trustworthy tool, grounded in theoretical axioms, for addressing challenges posed by black-box models like deep neural networks.
However, computing Shapley values encounters exponential complexity as the number of features increases.
Various approaches, including ApproSemivalue, KernelSHAP, and FastSHAP, have been explored to expedite the computation.
In our analysis of existing approaches, we observe that stochastic estimators can be unified as a linear transformation of randomly summed values from feature subsets.
Based on this, we investigate the possibility of designing simple amortized estimators and propose a straightforward and efficient one, \textbf{SimSHAP}, by eliminating redundant techniques.
Extensive experiments conducted on tabular and image datasets validate the effectiveness of our SimSHAP, which significantly accelerates the computation of accurate Shapley values.
\footnote{We will make our code publicly available once the paper has been accepted.}

\end{abstract}
\begin{figure}[htbp]
    \centering
    \vspace{-3mm}
    \begin{subfigure}{0.68\textwidth}
        \includegraphics[width=1\textwidth]{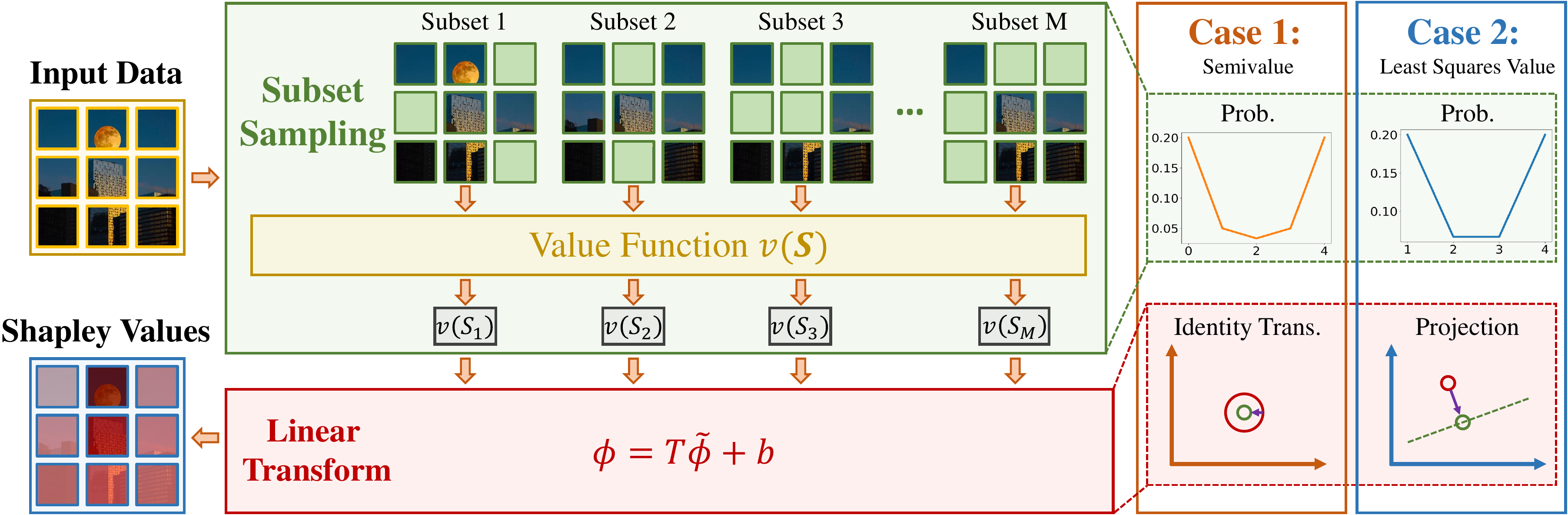}
        \caption{}
        \label{fig:framework_a}
    \end{subfigure}
    \hspace{2mm}
    \begin{subfigure}{0.29\textwidth}
        \includegraphics[width=1\textwidth]{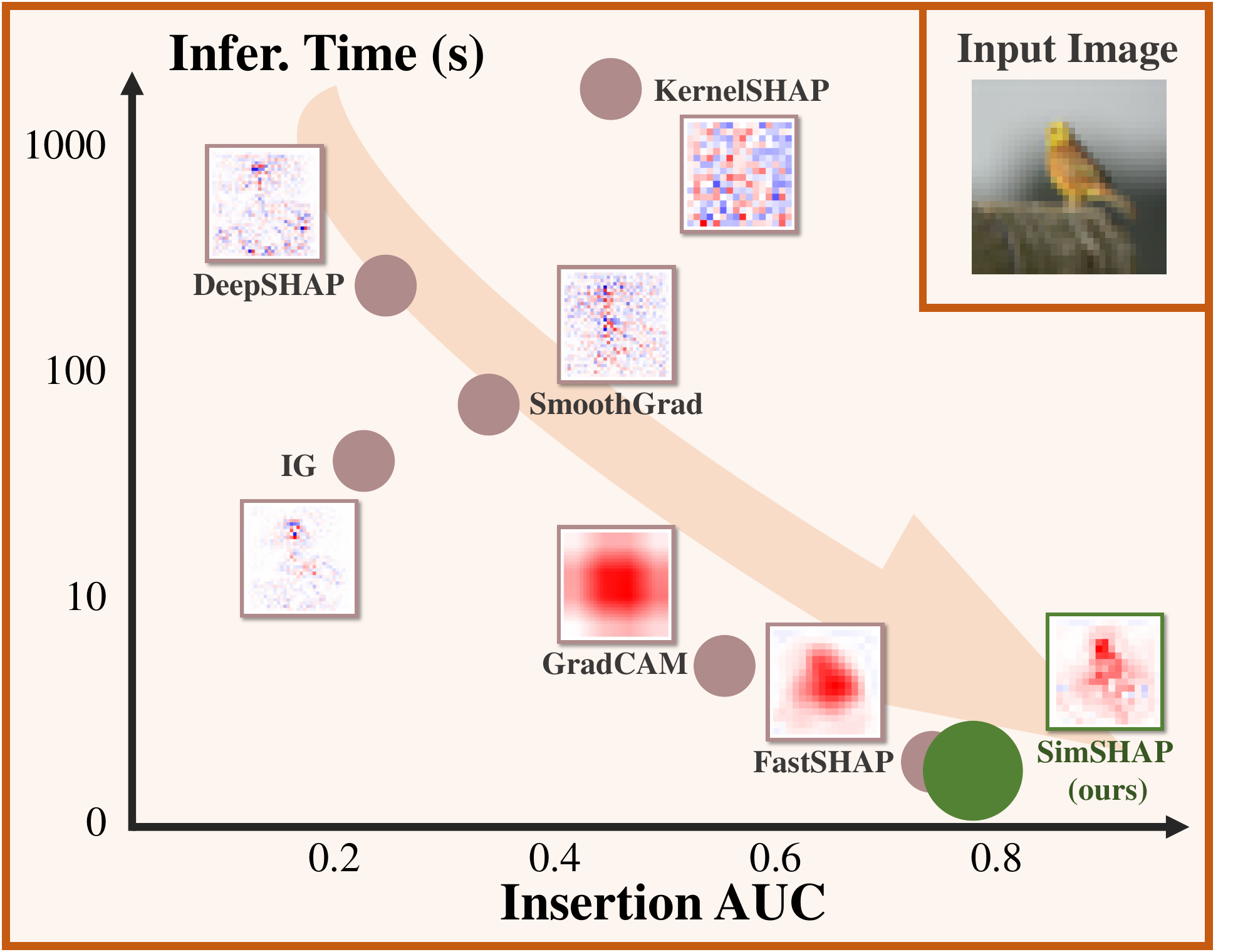}
        \caption{}
        \label{fig:framework_b}
    \end{subfigure}
    \vspace{-5mm}
    \caption{(a) Existing stochastic estimators for Shapley values can be unified as a linear transformation of the values obtained from sampled subsets. (b) We propose SimSHAP, which achieves high efficiency and maintain competitive approximation accuracy.}
\end{figure}

\vspace{-4mm}
\section{Introduction}
\vspace{-3mm}

Deep learning techniques have made significant strides in various industries by effectively and accurately learning complex functions. 
Despite their success, the lack of interpretability poses a challenge for the broader adoption of black-box models like deep neural networks in trust-demanding areas such as autonomous driving and healthcare. 
Explanability aims to establish a stable mapping from abstract representations to understandable concepts~\citep{gilpin2018explaining, zhang2021survey}. 
\textbf{Shapley values}, grounded on four fairness-based axioms (efficiency, symmetry, linearity, and dummy player), offer a stable and unique linear additive explanation~\citep{shapley1953value}. 
They achieve this by calculating the marginal contributions of each feature and combining them linearly. 
However, the computational demands of Shapley values can be prohibitive, especially for high-dimensional datasets~\citep{van2022tractability}.

Various methods have been proposed to expedite the computation of Shapley values in response to their high complexity, 
which are categorized into model-agnostic and model-specific methods~\citep{chen2023algorithms}.
\textbf{Model-agnostic methods}, such as semivalue~\citep{castro2009polynomial} and least squares value~\citep{lundberg2017unified, covert2021improving}, estimate Shapley values by sampling subsets of feature combinations.
\textbf{Model-specific methods} incorporate model-specific knowledge to speed up estimation.
For simple linear models, the computation cost can be reduced from exponential to linear~\citep{lundberg2017unified}.
Attempts have also been made to accelerate Shapley value computation for tree-based models~\citep{lundberg2020local} and neural networks~\citep{wang2021shapley, jethani2021fastshap, chen2023harsanyinet}.
Despite the significant advancements in accelerating Shapley value computation, distinguishing the precise variances between these algorithms remains challenging, thus complicating the selection of the most suitable algorithm for practical applications. 
Thus, investigating the interrelations among these algorithms is imperative.

This study examines Shapley value estimation methods, including semivalue ~\citep{castro2009polynomial} and least squares value~\citep{lundberg2017unified,covert2021improving}.
It is noted that these methods do not show significant differences.
Subsequently, a unified view on stochastic estimators is proposed, which involves linearly transforming the randomly summed values from feature subsets (see \cref{fig:framework_a}). 
Moreover, recent amortized estimators can also be unified as a fitting problem to Shapley values in different metric spaces.
In line with the principle of simplicity, we introduce \textbf{SimSHAP}, a simple and fast amortized Shapley value estimator. 
SimSHAP trains an amortized explanation model by minimizing the 
$l2$-distance to the (estimated) Shapley values in the Euclidean space. 
Compared to conventional methods~\citep{castro2009polynomial,lundberg2017unified}, SimSHAP achieves orders of magnitude faster computation with comparable accuracy as shown in \cref{fig:framework_b}.
Additionally, compared to recent amortized methods~\citep{jethani2021fastshap}, SimSHAP adopts an unconstrained optimization approach without subtle normalization~\citep{ruiz1998family}.
Extensive experiments on tabular and image datasets are conducted to showcase the effectiveness of SimSHAP.
Our key contributions are summarized as follows:
\vspace{-2mm}
\begin{itemize}
    \item \textbf{Unified view on Shapley value estimation.} 
    We unify various Shapley value estimation methods as the linear transformation of the randomly summed values from feature subsets.
    \vspace{-1mm}
    \item \textbf{SimSHAP, a simple and fast amortized estimator.}
    We propose SimSHAP, which minimizes the $l2$-distance to the estimated Shapley values within the Euclidean space.
    \vspace{-1mm}
    \item \textbf{Consistent efficiency improvement.}
    We show consistent efficiency improvement while maintaining accuracy through extensive experiments on both tabular and image datasets.
\end{itemize}
\vspace{-3mm}
\vspace{-2mm}
\section{Method}
\vspace{-3mm}

In this section, we start by explaining the symbols used in this paper and the basic knowledge of Shapley values in \cref{sec:background}.
Next, we introduce the unified view of Shapley value estimation especially for semivalue and least squares value in \cref{sec:strategy,sec:unified}.
Lastly, we propose SimSHAP, a simple and fast amortized Shapley value estimator in \cref{sec:simshap}.

\vspace{-1mm}
\subsection{Shapley Values} \label{sec:background}
\vspace{-2mm}

For a typical classification model $f:\mathcal{X} \mapsto \mathcal{Y}$, 
the input vector is denoted as $\vx = (\evx_1,\cdots,\evx_d) \in \mathcal{X}$ with $d$ dimensions, and $y\in \mathcal{Y} = \{1,\cdots,K\}$ represents the corresponding label.
Here, $\bm{S} \subset \bm{N} = \{1,\cdots,d\}$ refers to subsets of all feature indices, and $\bm{1}^{\bm{S}} \in \{0, 1\}^d$ is the corresponding indicator vector where $\bm{1}^{\bm{S}}_i = 1$ if $i \in \bm{S}$ and 0 otherwise.
Shapley values were initially introduced in cooperative game theory~\citep{shapley1953value} to ensure fair profit distribution among players. The formulation is as follows:
\begin{definition} \label{def:shapley}
    \textbf{(Shapley Values)}
    For any value function $v: P(\bm{N}) \mapsto \R$, where $P(\bm{N})$ is the power set of $\bm{N}$, the Shapley values $\phi(v) \in \R^d$ is computed by averaging the marginal contribution of each feature over all possible feature combinations as:
    \begin{align} \label{equ:shapley}
        \phi_i &= \sum_{\bm{S} \subset \bm{N} \setminus \{i\}} \frac{|\bm{S}|!(d-|\bm{S}|-1)!}{d!} \left( v(\bm{S} \cup \{i\}) - v(\bm{S}) \right), ~~i = 1,\cdots,d.
    \end{align}
\end{definition}
\begin{remark}
    Shapley values are the unique linear additive explanation that satisfies four fairness-based axioms (efficiency, symmetry, linearity, and dummy player)~\citep{shapley1953value}.
\end{remark}
When applying Shapley values, two considerations should be taken into account: (1) Machine learning models are not cooperative games; and (2) the complexity of \cref{equ:shapley} grows exponentially with the data dimension $d$.
Value function choices and estimation strategies are two key factors for applying Shapley values to machine learning models.
Numerous studies investigate the value function of Shapley values for machine learning models, examining the implications of the presence or absence of each feature~\citep{frye2021shapley,covert2021explaining,aas2021explaining}.
In this work, we adopt the classic value function $v(\bm{S}) = f(\bm{1}^{\bm{S}} \odot \vx)$ with masked input corresponding to the feature indices $\bm{S}$.
For a detailed overview of value function configurations, we suggest consulting additional work \citep{chen2023algorithms}.
We primarily focus on estimation strategies.

\vspace{-2mm}
\subsection{Estimation Strategy} \label{sec:strategy}
\vspace{-2mm}

To accelerate the computation of Shapley values, a practical approach is to estimate \cref{equ:shapley} by stochastic sampling.
Taking into account the specific characteristics of machine learning models, it is possible to decrease computational costs by merging or pruning redundant feature combinations.
In this section, we initially present three widely used stochastic estimators along with a recent amortized estimator, followed by an examination of the connections between them.

\vspace{-2mm}
\subsubsection{Stochastic Estimator} \label{sec:stochastic_esti}

\vspace{-3mm}
\paragraph{Semivalue}

In prior research, the classic formulation of Shapley values in \cref{equ:shapley} is named as \textbf{semivalue}~\citep{dubey1981value}.
The marginal contribution $v(\bm{S} \cup \{i\}) - v(\bm{S})$ represents the contribution of feature $i$ cooperated with subset $\bm{S}$.
Therefore, previous works, such as ApproSemivalue~\citep{castro2009polynomial}, propose to estimate Shapley values by Monte Carlo sampling as follows:
\vspace{-2mm}
\begin{align} \label{equ:semivalue}
    \phi^{sv}_i(v) &= \mathop{\mathbb{E}}_{\bm{S} \sim p^{sv}(\bm{S})} \left[ v(\bm{S} \cup \{i\}) - v(\bm{S}) \right]
    \approx \frac{1}{M} \sum_{k=1}^M \left[ v(\bm{S}_k \cup \{i\}) - v(\bm{S}_k) \right], 
\end{align}
where $p^{sv}(\bm{S}) = |\bm{S}|!(d-|\bm{S}|-1)!/d!$ is the sampling distribution, and $\bm{S}_k$ is the $k$-th sampled subset.

\vspace{-3mm}
\paragraph{Random Order Value}

Semivalue has an equivalent formulation named as \textbf{random order value}~\citep{shapley1953value,monderer2002variations}, which allocates credit to each feature by averaging contributions across all possible permutations.
Let $\pi: \{1,\cdots,d\} \mapsto \{1,\cdots,d\}$ be a permutation of feature indices, $\Pi(\bm{N})$ be the set of all permutations, and $H^i(\pi)$ be the set of predecessors of feature $i$ in permutation $\pi$.
The Shapley values are redefined and estimated as follows:
\vspace{-3mm}
\begin{align} \label{equ:rov}
    \phi^{ro}_i &= \mathop{\mathbb{E}}_{\pi \sim p^{ro}(\pi)} \left[ v(H^i(\pi) \cup \{i\}) - v(H^i(\pi)) \right] 
    \approx \frac{1}{M} \sum_{k=1}^M \left[ v(H^i(\pi) \cup \{i\}) - v(H^i(\pi)) \right],
\end{align}
where $p^{ro}(\pi) = 1/d!$ is the uniform distribution across all permutations.

\vspace{-3mm}
\paragraph{Least Squares Value}

Shapley values can be formulated in a linear regression fashion~\citep{ruiz1998family}.
Specifically, Shapley values $\bm{\phi}^{ls}$ are the minimum value of a constrained weighted least squares problem $\mathcal{L}(\bm{\phi})$ and we implement the KernelSHAP~\citep{lundberg2017unified, covert2021improving}
\footnote{See \cref{sec:unbiased} for unbiased KernelSHAP estimation suggested in a previous study~\citep{covert2021improving}.}
as follows:
\vspace{-1mm}
\begin{align} \label{equ:lsv}
    \min_{\bm{\phi}} \mathcal{L}(\bm{\phi})
    = \min_{\bm{\phi}}\sum_{\emptyset \subsetneq \bm{S} \subsetneq \bm{N}}
    \omega(\bm{S}) \left(v(\bm{S}) - v(\emptyset) - \bm{\phi}^T \bm{1}^{\bm{S}} \right)^2, 
    ~~s.t. ~ \bm{\phi}^T \bm{1}=v(\bm{N})-v(\emptyset)
\end{align}
where $\omega(\bm{S}) = \frac{d - 1}{\binom{d}{|\bm{S}|} |\bm{S}|(d-|\bm{S}|)}$ is also named as Shapley kernel~\citep{lundberg2017unified}.
Remarkably, the domain of $\omega(\bm{S})$ does not include the empty set $\emptyset$ and the full set $\bm{N}$.
\begin{proposition} \label{prop:equivalence}
    The least squares value in \cref{equ:lsv} is equivalent to the semivalue in \cref{equ:semivalue}.
\end{proposition}
\vspace{-1mm}
\begin{remark}
    Considering that the problem in \cref{equ:lsv} is convex, we can obtain the optimal solution strictly by Karush-Kuhn-Tucker (KKT) conditions~\citep{boyd2004convex}. See the proof of equivalence in \cref{sec:proof_equivalence}.
\end{remark}
\vspace{-2mm}
For high-dimension data, the optimization objective $\mathcal{L}$ in \cref{equ:lsv} requires an approximation:
\begin{align} \label{sec:kernelSHAP}
    \mathcal{L}(\bm{\phi}) &= \mathop{\mathbb{E}}_{\bm{S} \sim p^{ls}(\bm{S})} \gamma \left[ v(\bm{S}) - v(\emptyset) - \bm{\phi}^T \bm{1}^{\bm{S}} \right]^2
    \approx \frac{\gamma}{M} \sum_{k=1}^M \left[ v(\bm{S}_k) - v(\emptyset) - \eta^T \bm{1}^{\bm{S}_k} \right]^2,
\end{align}
where $\bm{S}_k$ is the $k$-th sampled subset, and $\gamma = \sum_{\emptyset \subsetneq \bm{S} \subsetneq \bm{N} } \omega(\bm{S})$, with $p^{ls}(\bm{S}) = \omega(\bm{S}) / \gamma$.

\vspace{-4mm}
\subsubsection{Amortized Estimator} \label{sec:amortized_esti}
\vspace{-3mm}

Stochastic estimators still encounter the trade-off between accuracy and efficiency.
To address this, it is necessary to gain insight into the internal characteristics of models.
For instance, in the case of linear models $f(\vx) = \omega^T \vx + b$ with value function $v(\bm{S}) = \omega^T (\bm{1}^{\bm{S}}\cdot \vx) + b$, 
the Shapley values can be easily computed in closed form as $\phi_i = \omega_i \vx_i$ with a linear complexity $\mathcal{O}(d)$.
Previous studies have also explored model-specific Shapley values estimators for various model structures, such as tree-based models~\citep{lundberg2020local} and neural networks~\citep{shrikumar2017learning,chen2018shapley,wang2021shapley,chen2023harsanyinet}.
However, these methods require subtle design and may even require special modules or training methodologies.
To tackle this, FastSHAP~\citep{jethani2021fastshap} trains an amortized parametric function $g(\vx; \theta): \mathcal{X} \mapsto \R^d$ to estimate Shapley values by penalizing predictions using a weighted least squares loss in \cref{equ:lsv}:
\vspace{-1mm}
\begin{align} \label{equ:fastshap}
    \theta = \arg \min_{\theta} 
    \mathop{\mathbb{E}}_{\vx \in \mathcal{X}}
    \mathop{\mathbb{E}}_{\bm{S} \sim p^{ls}(\bm{S})} 
    \left[ v(\bm{S}) - v(\emptyset) - g^T(\bm{S}; \theta) \bm{1}^{\bm{S}} \right]^2 
    ~~ s.t. ~ g^T(\vx; \theta) \bm{1} = v(\bm{N}) - v(\emptyset).
\end{align}
To optimize the constrained problem in \cref{equ:fastshap},
FastSHAP~\citep{jethani2021fastshap} adjusts predictions using additive efficient normalization~\citep{ruiz1998family} or optimizes with a penalty on the efficiency gap.

\vspace{-2mm}
\subsection{Unified Perspective} \label{sec:unified}
\vspace{-2mm}

\begin{table}[tbp] \small
    \centering
    \caption{Unified Stochastic Estimator \textcolor{gray}{\footnotesize{(``SV'' denotes Semivalue, ``LSV'' denotes Least Squares Value)}}.}
    \setlength\tabcolsep{4pt}
    \label{tab:unified_estimator}
    \begin{tabular}{l|ccccc}
        \toprule
        \textbf{Estimator} & \textbf{Prob. $p^i(\bm{S}$)} 
        & \textbf{Space $\Omega$}
        & \textbf{Coef. $a^i_{\bm{S}}$}
        & \textbf{Trans. $\mT$} & \textbf{Bias $\vb$} \\
        \midrule
        SV
        & $\left[ \binom{d}{|\bm{S}|} (|\bm{S}|\mathbb{I}_{i\in \bm{S}} + (d - |\bm{S}|)\mathbb{I}_{i\notin \bm{S}}) \right]^{-1}$
        & $\bm{S} \subset \bm{N}$
        & $2 \mathbb{I}_{i\in \bm{S}} - 1$
        & $\mI$ 
        & $\bm{0}$ \\
        LSV
        & $\left[ \gamma \binom{d}{|\bm{S}|} (d - |\bm{S}|) |\bm{S}| \right]^{-1}$
        & $\emptyset \subsetneq \bm{S} \subsetneq \bm{N} $
        & $\mathbb{I}_{i\in \bm{S}}$
        & $ \gamma (d\bm{I} - \bm{J})$ 
        & $\frac{v(\bm{N}) - v(\emptyset)}{d} \bm{1}$ \\
        \midrule
        \textbf{Ours}
        & $\left[ \gamma \binom{d}{|\bm{S}|} (d - |\bm{S}|) |\bm{S}| \right]^{-1}$
        & $\emptyset \subsetneq \bm{S} \subsetneq \bm{N} $
        & \makecell{
            $\gamma (d - |\bm{S}|) \mathbb{I}_{i \in \bm{S}}$ \\
            $- \gamma |\bm{S}| \mathbb{I}_{i \notin \bm{S}}$
        } 
        & $\mI$ 
        & $\frac{v(\bm{N}) - v(\emptyset)}{d} \bm{1}$ \\
        \bottomrule
    \end{tabular}
    \vspace{-5mm}
\end{table}

Numerous researches have been conducted to accelerate the stochastic estimation of Shapley values.
However, interconnections among these methods are not elucidated clearly.
This section proposes a comprehensive viewpoint on estimating Shapley values, 
which not only clarifies the distinct differences among these methods but also suggests a new direction for future research efforts.
\begin{definition} \label{def:unified_sampling}
    \textbf{Unified Stochastic Estimator}
    is defined as the linear transformation of the randomly summed values from feature subsets $\bm{S}$ as:
    \vspace{-3mm}
    \begin{align}
        \underbrace{\phi^{uni} = \mT \tilde{\phi} + \vb,}_{\text{The linear transformation}} ~~~
        \tilde{\phi}_i = \mathop{\mathbb{E}}_{\bm{S} \sim p^i(\bm{S})} \left[ a^i_{\bm{S}} v(\bm{S}) \right] 
        \underbrace{\approx \frac{1}{M} \sum_{k=1}^M a^i_{\bm{S}} v(\bm{S}_k)}_{\text{Monte Carlo sampling}},
    \end{align}
    where $\mT \in \R^{d\times d}$ and $\vb \in \R^d$ are transformation parameters, and $a^i_{\bm{S}}$ is the coefficient of subset $\bm{S}$.
\end{definition}
\begin{remark}
    When each subset $\bm{S}$ is sampled independently, the variance of the sampled $\tilde{\phi}_i$ is equal to $\mathbb{D}[a_{\bm{S}}^i v(\bm{S})] / M$, 
    indicating that a higher value of $M$ and lower fluctuation of $a_{\bm{S}}^i v(\bm{S})$ will result in a more stable estimation. 
    Various methods of Shapley value estimation are particular instances of \cref{def:unified_sampling}, as depicted in \cref{tab:unified_estimator}. 
    Further analysis is elaborated below. 
\end{remark}

\vspace{-2mm}
\paragraph{Semivalue}
The equivalence between the semivalue in \cref{equ:semivalue} and the random order value in \cref{equ:rov} is evident because the number of permutations is $(d - \vert \bm{S} \vert - 1)!\vert \bm{S} \vert!$ for a given subset $\bm{S}$ and feature $i$ positioned after features in $\bm{S}$. 
The semivalue is redefined within the framework of \cref{def:unified_sampling} as:
\vspace{-1mm}
\begin{align} \label{equ:uni_sv}
    \phi_i^{sv} &= \mT^{sv}_i \tilde{\phi}^{sv} + \vb^{sv}_i 
    = \sum_{\bm{S} \subset \bm{N}} p^i(\bm{S}) a^{sv}_{\bm{S}} v(\bm{S})
    = \underbrace{\sum_{\bm{S} \subset \bm{N} \setminus \{i\}} \frac{v(\bm{S} \cup \{i\}) - v(\bm{S})}{\binom{d}{|\bm{S}|}(d - |\bm{S}|)}}_{\text{The Shapley values}}.
\end{align}
The sampling probability $p^i(\bm{S})$ varies according to the index $i$. 
Therefore, this estimation method is not suitable for parallel computing devices like GPUs.

\vspace{-1mm}
\paragraph{Least Squares Value}
The least squares value is derived using the Lagrange multiplier method and is also presented as a unified form according to \cref{def:unified_sampling} as follows:
\begin{align} \label{equ:uni_lsv}
    \phi^{ls} &= \mT^{ls} \tilde{\phi}^{ls} + \vb^{ls} 
    = (d\mI - \mJ) \tilde{\phi}^{ls} + \frac{v(\bm{N}) - v(\emptyset)}{d} \bm{1} \\
    \tilde{\phi}^{ls}_i &= \sum_{\emptyset \subsetneq \bm{S} \subsetneq \bm{N} } p^i(\bm{S}) a^{ls}_{\bm{S}} v(\bm{S})
    = \sum_{\emptyset \subsetneq \bm{S} \subsetneq \bm{N} } \frac{v(\bm{S}) \mathbb{I}_{i\in \bm{S}}}{\binom{d}{|\bm{S}|} (d - |\bm{S}|) |\bm{S}|},
\end{align}
where $\bm{J}$ represents a matrix filled with ones. 
A detailed derivation can be found in \cref{sec:proof_equivalence}.
From \cref{equ:uni_lsv}, it is evident that the least squares value is essentially another form of direct sampling rather than the result of minimizing the least squares loss.

Model-agnostic stochastic estimators continue to face challenges in balancing accuracy and efficiency due to the inherent exponential complexity associated with potential feature subsets. 
A recent amortized estimator~\citep{jethani2021fastshap} accelerates the computation by leveraging the internal structure of the model, requiring only a single forward pass to estimate Shapley values.
Analogously to \cref{def:unified_sampling}, we derive the unified form of amortized estimators as follows:
\begin{definition} \label{def:unified_amortied}
    A \textbf{Unified Amortized Estimator} refers to a learnable parametric function $g(\vx;\theta): \mathcal{X} \mapsto \R^d$ used to fit true or estimated Shapley values $\bm{\phi}_{\vx}$ by minimizing the following loss function:
    \begin{align} \label{equ:unified_amortied}
        \theta = \arg \min_{\theta} \mathop{\mathbb{E}}_{\vx \in \mathcal{X}}
        \left[ \left\| g(\vx; \theta) - \bm{\phi}_{\vx} \right\|_{\mM}^2 \right],
    \end{align}
    where $\mM \in \sS^+$ is the metric matrix.
\end{definition}
To simplify notation, when given the non-empty proper subset order $(\bm{S}_1,\cdots,\bm{S}_n)$ with $n = 2^d - 2$,
we define the value vector $\vv = (v(\bm{S}_1), \cdots, v(\bm{S}_n)) \in \R^n$, $\vv_{\Delta} = \vv - v(\emptyset) \bm{1} \in \R^n$,
the indicator matrix $\mU = (\bm{1}^{\bm{S}_1}, \cdots, \bm{1}^{\bm{S}_n})^T \in \{0, 1\}^{n\times d}$,
and the weight matrix $\mW = diag(\omega(\bm{S}_1), \cdots, \omega(\bm{S}_n))$, with $\omega$ being the Shapley kernel. 
By setting $\bm{\phi}_{\vx}$ equal to or approximate the value of $(\mU^T \mW \mU)^{-1} \mU^T \mW \vv_{\Delta}$ and defining $\mM = \mU^T \mW \mU$, \cref{equ:unified_amortied} simplifies to the least squares loss of FastSHAP~\citep{jethani2021fastshap}, given by\footnote{For simplicity, we only present the case when $\phi_{\vx}$ is true value.}:
\begin{align} \label{equ:fastshap_equal}
    \mathcal{L} &= \mathop{\mathbb{E}}_{\vx \in \mathcal{X}}
    \left[ \left\Vert g(\vx;\theta) - (\mU^T \mW \mU)^{-1} \mU^T \mW \vv_{\Delta} \right\Vert^2_{\mU^T \mW \mU} \right]
    = \mathop{\mathbb{E}}_{\vx \in \mathcal{X}}
    \left[ \Vert \mU g(\vx;\theta) - \vv_{\Delta} \Vert_{\mW}^2 \right] + C.
\end{align}
Since the objective of FastSHAP is biased, the predictions from FastSHAP require correction through additive efficient normalization~\citep{ruiz1998family} given by:
$g(\vx;\theta) \leftarrow \frac{d\mI - \mJ}{d} g(\vx;\theta) + \frac{v(\bm{N}) - v(\emptyset)}{d} \bm{1}$.

\subsection{SimSHAP} \label{sec:simshap}

\begin{wrapfigure}[16]{r}{0.55\textwidth}
    \vspace{-6mm}
    \begin{algorithm}[H]
        \LinesNumbered
        \caption{SimSHAP training.}
        \KwIn{Value function $v$, learning rate $\alpha$, number of subsets $M$.}
        \KwOut{SimSHAP explainer $g(\vx;\theta)$.}
        \label{alg:simshap}
        \BlankLine
        Initialize random weights $\theta$; \\
        \While{not converged}{
            Sample the input $\vx \in \mathcal{X}$; \\
            Sample $M$ subsets $\{ \bm{S}_i | \bm{S}_i \sim p^{ls}(\bm{S}) \}$; \\
            Compute the estimated Shapley values as: 
            $\hat{\phi} = \frac{1}{M} \sum_{k=1}^M 
            \gamma ((d - |\bm{S}|)\mathbb{I}_{i\in \bm{S}} - |\bm{S}|\mathbb{I}_{i\notin \bm{S}}) v(\bm{S_k})
             + \frac{v(\bm{N}) - v(\emptyset)}{d} \bm{1}$; \\
            Compute 
            $\mathcal{L} = \Vert g(x;\theta) - \hat{\phi} \Vert_2^2$;\\
            Update the parameters as:
            $\theta \leftarrow \theta - \alpha \nabla_{\theta} \mathcal{L}$; \\
        }
    \end{algorithm}
\end{wrapfigure}

In \cref{sec:unified}, a new amortized estimator called \textbf{SimSHAP} is proposed for discussion.
Unlike FastSHAP's specialized metric matrix, we opt for the simple choice of using the identity matrix $\mM = \mI$.
Drawing on the strengths of semivalue and least squares value, a novel sampling approach is introduced, as detailed in the last row of \cref{tab:unified_estimator}.
This method involves sampling subsets $\bm{S}$ based on the distribution of least squares values $p^{ls}(\bm{S})$, which is independent of index $i$ and facilitates parallel computation.
Moreover, we define the coefficient as $a^i_{\bm{S}} = \gamma (d - |\bm{S}|) \mathbb{I}_{i \in \bm{S}} - \gamma |\bm{S}| \mathbb{I}_{i \notin \bm{S}}$.
The complete algorithm for SimSHAP is presented in \cref{alg:simshap}.

It is obvious that the expectation of the fitting target of SimSHAP is unbiased as:
\begin{align} \label{equ:unbias_simshap}
    \mathbb{E} [\phi_{\vx}] 
    &= \mT \sum_{\emptyset \subsetneq \bm{S} \subsetneq \bm{N} } p^{ls}(\bm{S}) a_{\bm{S}} v(\bm{S}) + \vb
    = \sum_{\emptyset \subsetneq \bm{S} \subsetneq \bm{N} } 
    \frac{[(d - |\bm{S}|)\mathbb{I}_{i\in \bm{S}} - |\bm{S}|\mathbb{I}_{i\notin \bm{S}}]v(\bm{S})}{\binom{d}{|\bm{S}|}(d - |\bm{S}|)|\bm{S}|} + \frac{v(\bm{N}) - v(\emptyset)}{d} \bm{1} \nonumber \\
    &= \sum_{\bm{S} \subset \bm{N} \setminus \{i\}} \frac{v(\bm{S} \cup \{i\}) - v(\bm{S})}{\binom{d}{|\bm{S}|}|\bm{S}|(d - |\bm{S}|)} \rightarrow \text{The Shapley values} .
\end{align}
If the capacity of $g(\vx;\theta)$ is large enough, SimSHAP will converge towards the true Shapley values with high accuracy.
The key distinctions between FastSHAP and SimSHAP are outlined in \cref{tab:unified_amortied_estimator}.
Both algorithms exhibit comparable time complexities (see detailed comparison in \cref{sec:algorithm_complexity}).

\begin{table}[tbp] \small
    \centering
    \caption{Unified Amortized Estimator.}
    \label{tab:unified_amortied_estimator}
    \begin{tabular}{l|ccc}
        \toprule
        \textbf{Estimator} & \textbf{$\mathbb{E} [\phi_{\vx}]$} 
        & \textbf{Metric Matrix $\mM$}
        & \textbf{Normalization} \\
        \midrule
        FastSHAP
        & $(\mU^T \mW \mU)^{-1}\mU^T \mW \vv_{\Delta}$
        & $\mU^T \mW \mU$
        & $\frac{d\mI - \mJ}{d}(\cdot) + \frac{v(\bm{N}) - v(\emptyset)}{d}\bm{1}$ \\
        \midrule
        \textbf{SimSHAP \scriptsize{(Ours)}}
        & $\frac{d\mI - \mJ}{d - 1} \mU^T \mW \vv + \frac{v(\bm{N}) - v(\emptyset)}{d} \bm{1}$
        & $\mI$
        & (unbiased) \\
        \bottomrule
    \end{tabular}
    \vspace{-5mm}
\end{table}

\section{Related Work}

\paragraph{Shapley Values Explanation}
Numerous studies~\citep{zhang2018visual,gilpin2018explaining,zhang2021survey,bodria2021benchmarking} have delved into elucidating the intricate workings of black-box models. 
To ensure fidelity, explanation methods should establish a stable mapping between abstract feature representations and concrete human concepts.
Shapley values originated in cooperative game theory~\citep{shapley1953value} to quantify each player's contribution within a coalition.
These values are unique in satisfying four fairness principles: efficiency, symmetry, linearity, and the dummy player concept~\citep{shapley1953value,dubey1981value,lipovetsky2001analysis}.
Recently, Shapley values have been adopted in machine learning to interpret predictions of black-box models like DNNs.
However, as machine learning models were not initially designed for cooperative games, 
defining value functions remains a challenge~\citep{covert2021explaining,frye2021shapley,aas2021explaining}.
Selecting suitable value functions is pivotal to prevent information leakage and ensure explanations align with human understanding~\citep{rong2022consistent,jethani2023don}.
Moreover, exact computation of Shapley values is often impractical due to the exponential data dimension complexity, leading to the development of efficient estimator strategies divided as \textbf{model-agnostic} and \textbf{model-specific} methods.

\paragraph{Model-Agnostic Estimation}
Shapley values are a linear transformation of values across all feature combinations. 
To address the exponential complexity of exact computations, various methods have been proposed to estimate Shapley values by sampling potential combinations. 
\textbf{Semivalues}~\citep{dubey1981value} represent the classic form of Shapley values, 
while ApproSemivalue~\citep{castro2009polynomial} offers a polynomial-time approximation using Monte Carlo sampling.
\textbf{Random order values}~\citep{monderer2002variations} reframe Shapley values as the expectation of marginal contributions with random permutations, efficiently estimated by uniformly sampling feature permutations~\citep{castro2009polynomial,strumbelj2010efficient}. 
Additionally, \textbf{least squares values}~\citep{charnes1988extremal,ruiz1998family,lipovetsky2001analysis} reveal Shapley values as solutions to least squares problems. 
KernelSHAP~\citep{lundberg2017unified,covert2021improving} leverages this property to estimate Shapley values through linear regression on sampled subsets.
Given the myriad proposed strategies to expedite Shapley value estimation, exploring differences between these methods becomes pertinent. 
Hence, this study delves into a unified stochastic estimator.

\paragraph{Model-Specific Estimation}
To enhance the efficiency of Shapley value estimation further, some approaches trim redundant subsets by leveraging specific machine learning model structures. 
For linear models, Shapley value computation complexity can be reduced to linear time due to model linearity~\citep{lundberg2017unified}. 
TreeExplainer~\citep{lundberg2020local} computes local explanations for tree-based models based on exact Shapley values in polynomial time.
For deep neural networks, DeepLIFT~\citep{shrikumar2017learning} efficiently computes attributions by backpropagating neuron contributions to input features. 
ShapleyNet~\citep{wang2021shapley} and HarsanyiNet~\citep{chen2023harsanyinet} introduce specialized network structures for Shapley value computation in a single pass. 
Certain methods~\citep{chen2018lshapley,teneggi2022fast} achieve exact or finite-sample Shapley value approximations by assuming distributions, 
albeit extending these methods to diverse network structures and distributions poses challenges.
Recent efforts~\citep{schwarzenberg2021efficient,jethani2021fastshap,chuang2023cortx} introduce learnable parametric functions to approximate explanations, aiming to enhance flexibility. 
Notably, amortized methods can be unified, leading to the proposal of SimSHAP.
\section{Experiments}

\subsection{Structured Data Experiments}

The performance of SimSHAP is evaluated by comparing it with popular baseline methods. 
Initially, we analyze its accuracy and reliability using various tabular datasets, comparing model predictions with actual Shapley values. 
Subsequently, we assess image interpretation with conventional metrics using a well-known image dataset.
All experiments are done with a 3090 GPU card.

\subsubsection{Experimental Details} \label{sec:detail_1}

\paragraph{Dataset Description} \label{sec:dataset_1}
The accuracy of SimSHAP is tested on different tabular datasets including \texttt{census}, \texttt{news}, and \texttt{bankruptcy} for thorough evaluation. 
The \texttt{census} dataset is derived from the 1994 United States Census database, comprising 12 features with labels indicating if an individual's income exceeds \$50,000/year~\citep{kohavi1996scaling}. 
The \texttt{news} dataset consists of 60 numerical features related to articles on Mashable over two years, with the label showing if the article's share count exceeds the median value of 1400 \citep{fernandes2015proactive}. 
The \texttt{bankruptcy} dataset is sourced from the Taiwan Economic Journal from 1999 to 2009, containing 96 features with the label indicating whether a company went bankrupt \citep{liang2016financial}. 
These datasets are divided into 80/10/10 splits for training, validation, and testing purposes.

\paragraph{Implementation Details} \label{sec:impl_1}
For the original models $f:\mathcal{X} \mapsto \mathcal{Y}$, tree-based methods are chosen for all datasets.
Following the strategy of FastSHAP \citep{jethani2021fastshap}, neural networks are trained as surrogate models and utilized as the value function for training the explanation model. 
The SimSHAP explainer model $\phi_{fast}(\bm{x}, \bm{y}; \theta)$ is implemented as an MLP $g(\bm{x},\theta): \mathcal{X}\mapsto \mathbb{R} ^d\times \mathcal{Y}$, 
generating a vector of Shapley values for each $y\in \mathcal{Y}$ in accordance with FastSHAP's methodology \citep{jethani2021fastshap}. 
To balance the speed and accuracy of training, 64 samples per data $\bm{x}$ are chosen, pair sampling is employed, and each tabular dataset is trained for 1000 epochs. 
More details on our code and hyperparameters can be found in our codebase and \cref{sec:details}.
For a comprehensive evaluation, SimSHAP is compared against various baselines, predominantly non-amortized iterative methods. 
Specifically, comparisons are made with KernelSHAP \citep{lundberg2017unified} and its enhanced version using pair sampling \citep{covert2021improving}. 
We also evaluate against permutation sampling and an enhanced method using antithetical sampling \citep{mitchell2022sampling}. 
Additionally, FastSHAP is included in the comparison. 
To compute the distance, KernelSHAP is run until convergence for a specified threshold to compute the ground truth Shapley values. 

\begin{figure}
  \centering
  \includegraphics[width=0.8\textwidth]{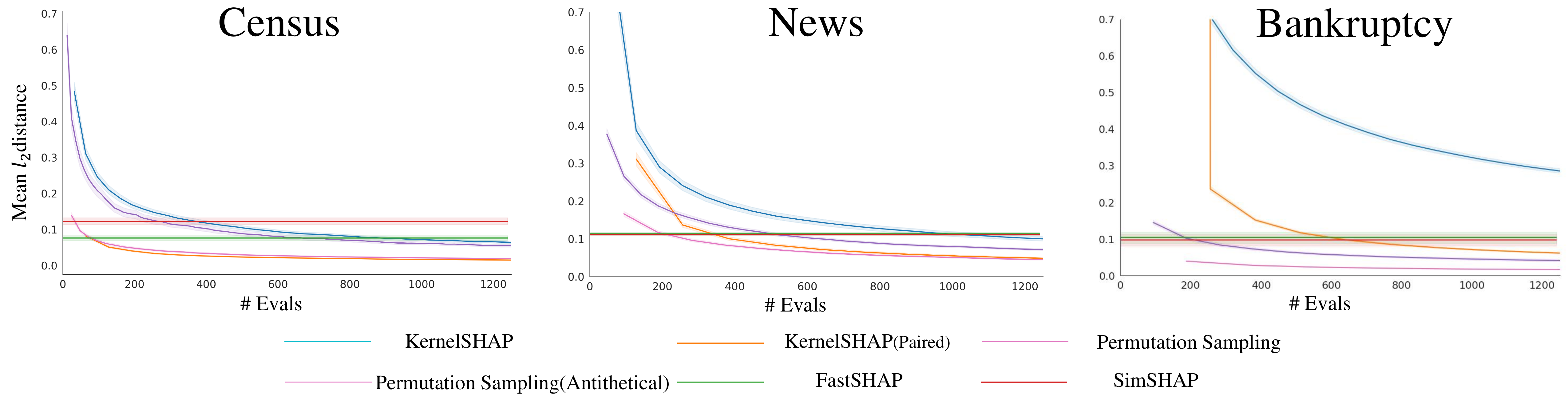}
  \caption{Accuracy of SimSHAP estimation across tabular datasets.}
  \label{fig:losscurve}
\end{figure}

\subsubsection{Quantitative Experiments}

For quantitative evaluations, we have selected three datasets. 
With access to ground truth Shapley values for tabular datasets, we can compare the accuracy of various methods. 
To assess the differences, we use $l_1$ and $l_2$ distances between estimations and ground truth.
In \cref{fig:losscurve}, we juxtapose SimSHAP against non-amortized methods and FastSHAP. 
All explanation methods are evaluated using the same surrogate model. 
The results indicate that SimSHAP achieves comparable accuracy to FastSHAP across diverse datasets. 
Notably, pair sampling, found to be more effective in two KernelSHAP estimations in non-amortized methods, has been integrated into SimSHAP.
Furthermore, we examine this matter through formula analysis, noting that the weights of both \textbf{Semivalue} and \textbf{Least Square Value} are symmetric, favoring smaller and larger values. 
Additionally, antithetical sampling has proven to be more effective than the Monte Carlo permutation method.
Moreover, we conduct experiments to investigate the impact of different hyperparameters on the accuracy and convergence rate of SimSHAP. 
For further details, please see \cref{sec:ablation}.

\subsection{Image Data Experiments}

Addressing the complex and high-dimensional nature of image data poses significant challenges for interpretation techniques. 
This section evaluates the effectiveness of SimSHAP on the CIFAR-10~\citep{krizhevsky2009learning} dataset by comparing it with commonly used baseline methods.
All experiments are done with a 3090 GPU card.

\subsubsection{Experimental Details} \label{sec:detail_2}

\paragraph{Dataset Description} \label{sec:dataset_2}
The explainability of SimSHAP is tested on the \textbf{CIFAR-10} dataset, 
which comprises 60000 images of dimensions $32\times 32$ categorized into 10 classes. 
In our experiments, 50000 images are used for training, while the remaining 10000 are equally divided for validation and testing purposes.

\paragraph{Implementation Details} \label{sec:impl_2}
Both the original model and surrogate models are structured based on ResNet-18. 
Following a similar approach to FastSHAP, image attributions are assigned to $16\times 16$ regions, where each element corresponds to a $2\times 2$ superpixel. 
The SimSHAP explainer model is implemented using a U-net~\citep{ronneberger2015u} structure. 
In comparison to various Shapley value estimators, our SimSHAP method is evaluated. 
KernelSHAP and DeepSHAP~\citep{lundberg2017unified}, based on the Shapley Additive Explanation principle, offer a unified measure of feature importance. 
While KernelSHAP is model-agnostic, DeepSHAP is tailored for neural networks to enhance performance. 
Additionally, we incorporate the KernelSHAP-S method, similar to FastSHAP~\citep{jethani2021fastshap}, which employs a surrogate model as the value function.
Furthermore, SimSHAP is contrasted with other gradient-based methods such as Integrated Gradients~\citep{sundararajan2017axiomatic}, SmoothGrad~\citep{smilkov2017smoothgrad}, and GradCAM~\citep{selvaraju2017grad}. 
In our SimSHAP implementation, 8 samples per image are used, pair sampling is employed, and the training duration spans 500 epochs.

\subsubsection{Qualitative Experiments}

\begin{figure}[tbp]
  \centering
  \includegraphics[width=0.9\textwidth]{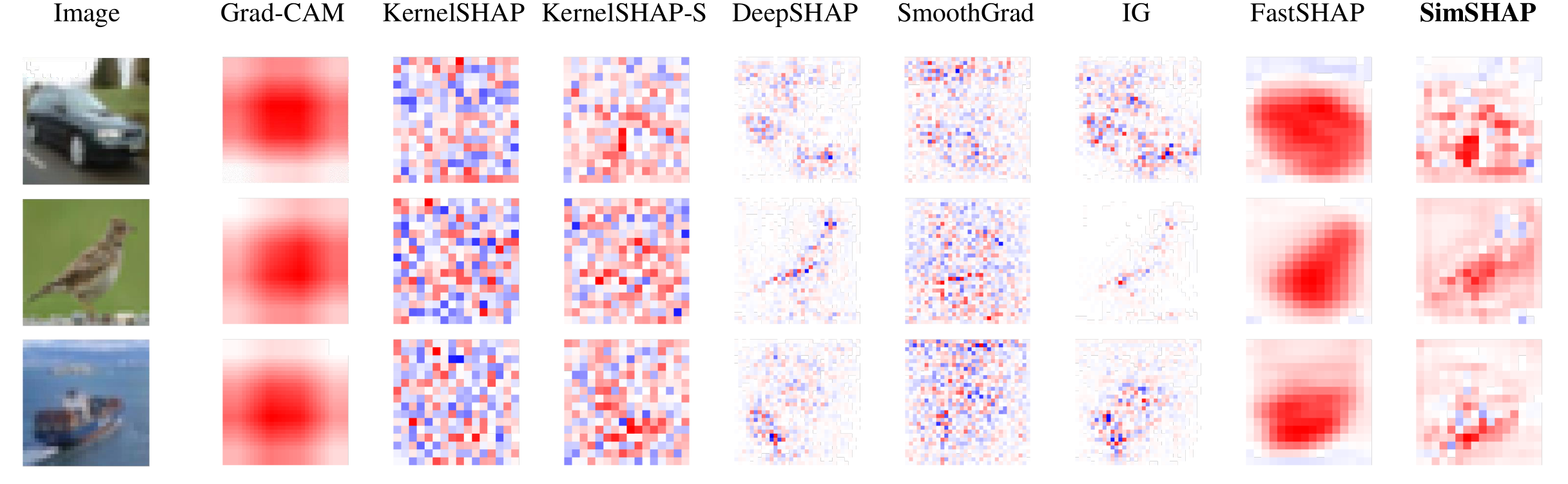}
  \caption{Comparison of different methods on randomly-chosen images in CIFAR-10.}
  \label{fig:comparison}
\end{figure}
\begin{figure}[tbp]
  \centering
  \includegraphics[width=0.7\textwidth]{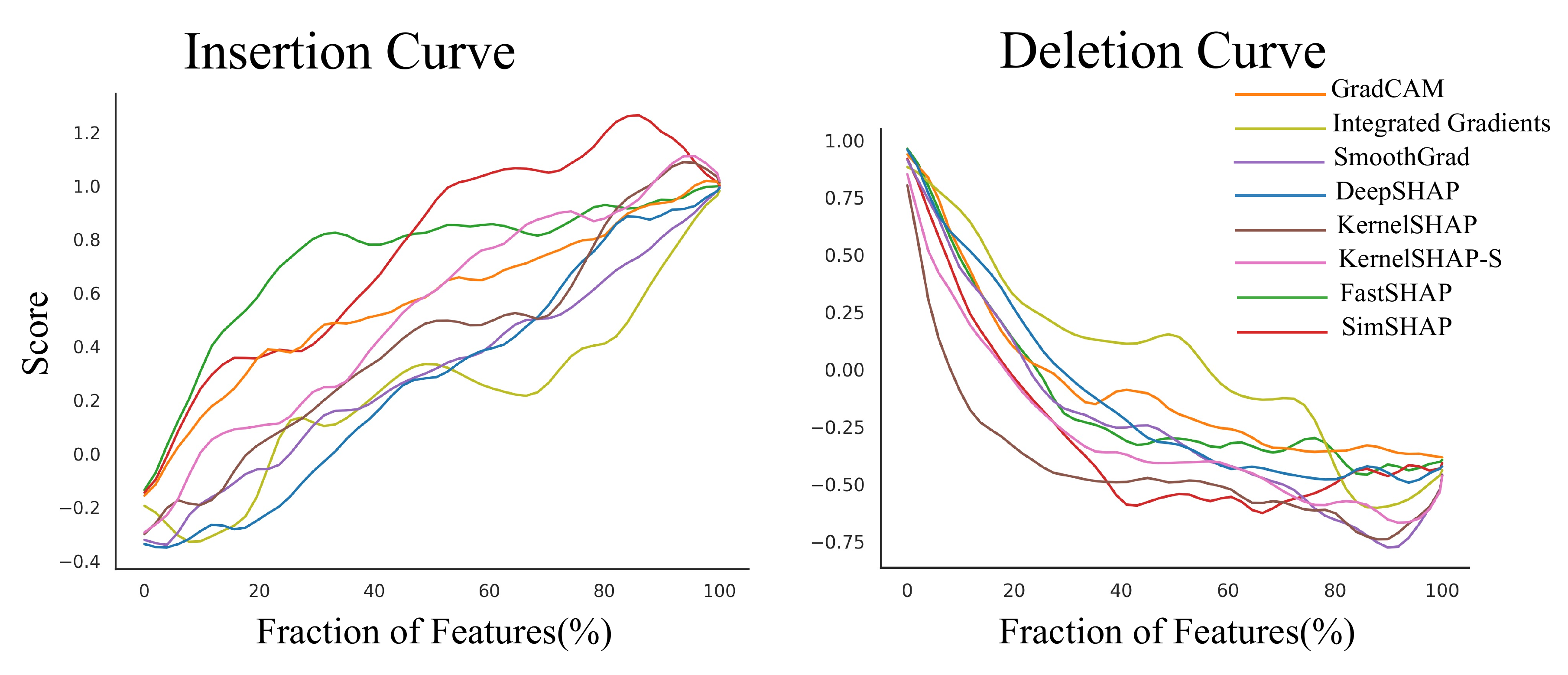}
  \caption{Mean Insertion and Deletion score curves for different methods on CIFAR-10 dataset}
  \label{fig:metric}
\end{figure}

We begin by showcasing three randomly selected images from the test set along with their respective attributions generated by various methods, as depicted in \cref{fig:comparison}. 
SimSHAP, similar to GradCAM and FastSHAP, is adept at identifying crucial regions but excels in outlining the contours of the primary object. 
One key advantage of SimSHAP over DeepSHAP or Integrated Gradients is its ability to assign varying degrees of importance to different parts of the object. 
For instance, the SimSHAP interpretation in the second row reveals that the model emphasizes the body and tail of the bird rather than the head when classifying the image as "Bird".
In contrast, KernelSHAP and KernelSHAP-S produce somewhat chaotic attributions, while SmoothGrad struggles to pinpoint the main object.
Based on a qualitative assessment, it can be inferred that SimSHAP holds promise as an effective method for image explanation. 
Subsequently, we will present quantitative experiments to further substantiate this claim.

\subsubsection{Quantitative Experiments}

Evaluating the accuracy of image explanations is challenging due to the computational complexity associated with high-dimensional data. 
To address this, we utilize the \textbf{Insertion / Deletion metrics}~\citep{petsiuk2018rise}, which assess how well explanations pinpoint informative regions within the image. 
The \textit{Deletion} metric examines how the model's decision evolves as crucial pixels are progressively removed, 
while the \textit{Insertion} metric follows the reverse path. 
These metrics gather scores based on the target class output of the masked image. 
Throughout the masking process, scores are normalized so that they start with 0/1 and end with 1/0 for Insertion/Deletion metrics.
To calculate these metrics, we sequentially delete or insert pixels in the order of attributions, plot the resulting curve, and determine the area under the curve (AUC). 
Following the guidance of \cite{petsiuk2018rise}, we establish the baseline image as zero for deletions and employ a Gaussian-blurred image for insertions.

\cref{fig:metric} illustrates the mean insertion score and deletion score curves. 
In terms of insertion, SimSHAP outperforms FastSHAP by achieving the highest values towards the end of the curve. 
Concerning deletion, SimSHAP demonstrates results similar to KernelSHAP-S, with a relatively rapid decrease and consistently low scores.

\begin{wraptable}[25]{r}{0.55\textwidth} \small
    \centering
    \caption{Insertion / Deletion metrics on CIFAR-10.}
    \tabcolsep=0.1cm
    \begin{tabular}{l|cccccc}
    \toprule
    \multirow{2}[4]{*}{} & \multicolumn{6}{c}{\textbf{CIFAR-10}} \\
\cmidrule{2-7}          & \multicolumn{3}{c}{\textbf{Insertion AUC↑}} & \multicolumn{3}{c}{\textbf{Deletion AUC↓}} \\
    \midrule
    FastSHAP & \multicolumn{3}{c}{\textbf{0.748 \scriptsize{(±0.082)}}} & \multicolumn{3}{c}{-0.133 \scriptsize{(±0.055)}} \\
    GradCAM & \multicolumn{3}{c}{0.563 \scriptsize{(±0.044)}} & \multicolumn{3}{c}{-0.075 \scriptsize{(±0.034)}} \\
    IG & \multicolumn{3}{c}{0.241 \scriptsize{(±0.051)}} & \multicolumn{3}{c}{0.033 \scriptsize{(±0.052)}} \\
    SmoothGrad & \multicolumn{3}{c}{0.318\scriptsize{(±0.052)}} & \multicolumn{3}{c}{-0.246 \scriptsize{(±0.103)}} \\
    DeepSHAP & \multicolumn{3}{c}{0.291 \scriptsize{(±0.101)}} & \multicolumn{3}{c}{-0.140 \scriptsize{(±0.173)}} \\
    KernelSHAP & \multicolumn{3}{c}{0.430 \scriptsize{(±0.064)}} & \multicolumn{3}{c}{\textcolor[rgb]{ 1,  0,  0}{\textbf{-0.443 \scriptsize{(±0.157)}}}} \\
    KernelSHAP-S & \multicolumn{3}{c}{0.542 \scriptsize{(±0.052)}} & \multicolumn{3}{c}{\textbf{-0.305 \scriptsize{(±0.152)}}} \\
    \textbf{SimSHAP (Ours)} & \multicolumn{3}{c}{\textcolor[rgb]{ 1,  0,  0}{\textbf{0.757 \scriptsize{(±0.117)}}}} & \multicolumn{3}{c}{-0.302 \scriptsize{(±0.063)}} \\
    \bottomrule
    \end{tabular}%
  \label{tab:del-ins}%
    \vspace{4mm}
    \caption{Inference time(s) / training time(min) on both tabular and image datasets.}
    \begin{tabular}{ll|cccc}
    \toprule
    & \multicolumn{1}{c|}{} & \textbf{Census} & \textbf{News} & \textbf{Bank} & \textbf{CIFAR-10} \\
    \cmidrule{1-6}    
    \multirow{6}[2]{*}{\rotatebox{90}{\textbf{Inference}}} 
    & FastSHAP & 0.004  & 0.004  & 0.005  & 0.090  \\
    & GradCAM & -     & -     & -     & 1.907  \\
    & IG & -     & -     & -     & 25.486  \\
    & KernelSHAP & 4.438  & 66.410  & 96.615  & 1440.595  \\
    & KernelSHAP-S & 43.560  & 67.002  & 101.053  & 1418.292  \\
    & \textbf{SimSHAP (Ours)} & \textcolor[rgb]{ 1,  0,  0}{\textbf{0.002}} & \textcolor[rgb]{ 1,  0,  0}{\textbf{0.001}} & \textcolor[rgb]{ 1,  0,  0}{\textbf{0.001}} & \textcolor[rgb]{ 1,  0,  0}{\textbf{0.086}} \\
    \midrule
    \multirow{2}[2]{*}{\rotatebox{90}{\textbf{Train}}} 
    & FastSHAP & 11.098  & 16.223  & 2.005  & \textbf{97.548} \\
    & \textbf{SimSHAP (Ours)} & \textbf{7.466} & \textbf{7.603} & \textbf{1.206} & 324.100 \\
    \bottomrule
    \label{tab:speed}
    \end{tabular}
\end{wraptable}%

Following the protocol~\citep{jethani2021fastshap}, a detailed comparison is presented in \cref{tab:del-ins}. 
SimSHAP displays the highest insertion AUC and the second-highest deletion AUC. 
While FastSHAP performs similarly to SimSHAP in terms of insertion AUC, 
GradCAM and KernelSHAP-S also show effectiveness. 
KernelSHAP excels in deletion AUC compared to all other methods, albeit with a notable standard deviation. 
Considering that a dependable Shapley value estimator should excel in both metrics, SimSHAP emerges as a compelling option.

\vspace{-1mm}
\subsubsection{Speed Evaluation}
\vspace{-1mm}

This section provides an analysis of the training and inference speeds of various methods applied to tabular and image datasets (refer to \cref{tab:speed} for details). 
Specifically, the evaluation focuses on SimSHAP and FastSHAP in terms of training speed.
The findings indicate that SimSHAP exhibits quicker training time (albeit requiring more epochs) compared to FastSHAP to achieve similar accuracy levels on tabular datasets. 
However, for image datasets, SimSHAP demands more time due to the larger number of required masks.
In terms of inference speed on tabular datasets, gradient-based methods are not suitable when tree-based models are utilized as the base model. 
SimSHAP demonstrates superior performance in inference speed over all baseline methods, being marginally faster than FastSHAP, as it eliminates the need to normalize the output of the explainer model for efficiency. 
On the other hand, gradient-based methods show slightly slower performance, while KernelSHAP and KernelSHAP-S require more time due to backward computation and numerous iterations during inference. 
Furthermore, the robustness of SimSHAP under limited data conditions is also tested. 
These experiments indicate that satisfactory performance can be attained with just 20\% samples. 
Additionally, exploratory experiments are conducted to study the impact of different hyperparameters on the accuracy and convergence rate of SimSHAP, including the number of samples, epochs, and learning rate choices. 
For further details, refer to \cref{sec:ablation}.
\vspace{-2mm}
\section{Limitations} \label{sec:limitation}
\vspace{-2mm}

This study presents a unified view on Shapley value estimation and introduces a simple amortized estimator called SimSHAP. 
However, there are certain limitations to this research. 
\textbf{Firstly}, as per the definition in \cref{def:unified_sampling}, the sampling stability of the unified stochastic estimator is directly related to the number of samples and the target function $a_{\bm{S}}^i v(\bm{S} )$. 
We have not delved into whether the unified estimator possesses an optimally theoretical sampling strategy. 
\textbf{Secondly}, in \cref{def:unified_amortied}, we introduce a metric matrix $\mM$ and opt to set it as the identity matrix $\mI$ due to experimental observations suggesting that the choice of $\mM$ has minimal impact on performance as long as it is (semi-)positive definite. 
We intend to further investigate the optimal selection of $\mM$.

\vspace{-2mm}
\section{Conclusion}
\vspace{-2mm}

The interpretability of black box models plays a vital role in enhancing model performance and establishing user confidence. 
Shapley values offer a dependable and interpretable attribution approach grounded in axiomatic principles. 
However, the computational complexity associated with Shapley values poses challenges to their practical utility. 
This study aims to elucidate the inherent relationships between existing stochastic estimators and the latest amortized estimators, 
presenting a unified perspective on estimation. 
In this context, we introduce SimSHAP as a simple and effective estimator. 
We also recognize that there are unresolved issues within this study as indicated in \cref{sec:limitation}, 
We look forward to future studies that will delve deeper into these aspects.

{
\small
\bibliographystyle{plainnat}
\bibliography{ref}
}


\appendix
\clearpage
\appendix
\section{Appendix}

\subsection{Proof of Equivalence between Semivalue and Least Squares Value} \label{sec:proof_equivalence}

This proposition was first carefully studied by~\cite{charnes1988extremal}, and here we review it.
\begin{proof}
    Before starting the proof, we investigate the characteristics of coefficients $\omega(\bm{S}) = \frac{d - 1}{\binom{d}{|\bm{S}|} |\bm{S}|(d-|\bm{S}|)}$ and find the equations as follows:
    \begin{align} \label{equ:lemma_omega}
        \sum_{\emptyset \subsetneq \bm{S} \subsetneq \bm{N}, i \in \bm{S}} 
        &\omega(\bm{S}) \phi^T \bm{1}^{\bm{S}} = A \phi_i + B \sum_{j\neq i} \phi_j,
        &\text{where}~ A = \frac{d - 1}{d} \sum_{i=1}^{d-1} \frac{1}{d-i}, ~B = A - \frac{d-1}{d}.
    \end{align}
    We follow the matrix definitions used in the main text here.
    given the order of non-empty proper subset $(\bm{S}_1,\cdots,\bm{S}_n)$ and $n = 2^d - 2$, 
    we define the value vector $\vv = (v(\bm{S}_1), \cdots, v(\bm{S}_n)) \in \R^n$,
    the indicator matrix $\mU = (\bm{1}^{\bm{S}_1}, \cdots, \bm{1}^{\bm{S}_n})^T \in \{0, 1\}^{n\times d}$,
    and the weight matrix $\mW = diag(\omega(\bm{S}_1), \cdots, \omega(\bm{S}_n))$, where $\omega$ is the Shapley kernel.
    We find that the matrix $\mU, \mW$ has some very nice properties according to \cref{equ:lemma_omega}:
    \begin{align}
        \mU^T \mW \mU &= \frac{d - 1}{d} \mI + B \mJ, ~
        (\mU^T \mW \mU)^{-1} &= \frac{d}{d - 1} \mI - C\mJ, ~\text{where} ~C = \frac{d^2 B}{(d - 1)(d^2 B + d - 1)} \nonumber
    \end{align}
    Considering that the problem in \cref{equ:lsv} is convex, we can obtain the optimal solution by Karush-Kuhn-Tucker (KKT) conditions~\citep{boyd2004convex}.
    We first formulate the Lagrangian function as:
    \begin{align} \label{equ:shap_lagrangian_matrix}
        L(\lambda, \bm{\phi}) &= \Vert \mU \bm{\phi} - \vv_{\Delta} \Vert_{\mW}^2 
        + \lambda (\bm{1}^T \bm{\phi} - v_{all}),
    \end{align}
    where $\vv_{\Delta} = \vv - v(\emptyset) \bm{1} \in \R^n, v_{all} = v(\bm{N}) - v(\emptyset) \in R$.
    Differentiating \cref{equ:shap_lagrangian_matrix} and setting the derivative to 0, we obtain:
    \begin{align}
        \nabla_{\bm{\phi}} L &= 2 \mU^T \mW (\mU \bm{\phi} - \vv_{\Delta}) + \lambda \bm{1} = 0 \label{equ:matrix_22} \\
        \nabla_{\lambda} L &= \bm{1}^T \bm{\phi} - v_{all} = 0 \label{equ:matrix_23}
    \end{align}
    After multiplying both sides of \cref{equ:matrix_22} by an all one vector $\bm{1}$,
    we can get:
    \begin{align}
        2 \bm{1}^T \mU^T \mW \mU \bm{\phi} - 2 \bm{1}^T \mU^T \mW \vv_{\Delta} + \lambda \bm{1}^T \bm{1} &= 0 \notag \\
        2 \bm{1}^T \left( \frac{d - 1}{d} \mI + B \mJ \right) \bm{\phi} - 2 \bm{1}^T \mU^T \mW \vv_{\Delta} + \lambda d &= 0 \notag \\
        2\left( \frac{d-1}{d} + Bd \right) \bm{1}^T \bm{\phi} - 2 \bm{1}^T \mU^T \mW \vv_{\Delta} + \lambda d &= 0 \notag \\
        2\left( \frac{d-1}{d} + Bd \right) v_{all} - 2 \bm{1}^T \mU^T \mW \vv_{\Delta} + \lambda d &= 0 \notag \\
        \frac{2}{d} \left[ \bm{1}^T \mU^T \mW \vv_{\Delta} - \left( \frac{d-1}{d} + Bd \right) v_{all} \right] &= \lambda. \label{equ:matrix_24}
    \end{align}
    Substituting \cref{equ:matrix_24} into \cref{equ:matrix_22} yields:
    \begin{align}
        2 \mU^T \mW \mU \bm{\phi} - 2 \mU^T \mW \vv_{\Delta} + \frac{2}{d} \left[ \bm{1}^T \mU^T \mW \vv_{\Delta} - \left( \frac{d-1}{d} + Bd \right) v_{all} \right] \bm{1} &= 0 \notag \\
        \left( \frac{d-1}{d} \bm{\phi} + \textcolor{blue}{B v_{all} \bm{1}} \right)
        - \mU^T \mW \vv_{\Delta} + \frac{1}{d} \mJ \mU^T \mW \vv_{\Delta} - 
        \left( \frac{d-1}{d^2} v_{all} \bm{1} + \textcolor{blue}{B v_{all} \bm{1}} \right) &= 0 \notag \\
        \frac{d-1}{d} \bm{\phi} - \mU^T \mW \vv_{\Delta} + \frac{1}{d} \mJ \mU^T \mW \vv_{\Delta} - \frac{d-1}{d^2} v_{all} \bm{1} &= 0 \notag \\
        \bm{\phi} = \frac{d\mI - \mJ}{d - 1} \mU^T \mW \vv_{\Delta} + \frac{v(\bm{N}) - v(\emptyset)}{d} &\bm{1} \notag \\
        \bm{\phi} = \left( \frac{d\mI - \mJ}{d - 1} \mU^T \mW \vv - 
        \underbrace{\textcolor{blue}{\frac{d\mI - \mJ}{d - 1} \mU^T \mW v(\emptyset) \bm{1}}}_{\text{Equals to 0}} \right) 
        + \frac{v(\bm{N}) - v(\emptyset)}{d} &\bm{1} \notag \\
        \bm{\phi} = \frac{d\mI - \mJ}{d - 1} \mU^T \mW \vv + \frac{v(\bm{N}) - v(\emptyset)}{d} &\bm{1}. \label{equ:matrix_shap_compute}
    \end{align}
    For parallel computing, we can rewrite $\mU^T \mW \vv$ in a sampling form as follows:
    \begin{align}
        \mU^T \mW \vv &= \sum_{\emptyset \subsetneq \bm{S} \subsetneq \bm{N}} \omega(\bm{S}) v(\bm{S}) \bm{1}^{\bm{S}} = \mathop{\mathbb{E}}_{\bm{S} \sim p^{ls}(\bm{S})} \gamma v(\bm{S}) \bm{1}^{\bm{S}} \\
        \text{where}~ \gamma &= \sum_{\emptyset \subsetneq \bm{S} \subsetneq \bm{N}} \omega(\bm{S}), ~
        p^{ls}(\bm{S}) = \omega(\bm{S}) / \gamma \notag
    \end{align}
    Next, we explain that \cref{equ:matrix_shap_compute} is equivalent to the Shapley value.
    For the $i$-th element of $\bm{\phi}$, we have:
    \begin{align}
        \phi_i &= \left( \frac{d}{d - 1} [\mU^T \mW \vv]_i - \frac{1}{d - 1} \bm{1}^T \mU^T \mW \vv \right) + \frac{v(\bm{N}) - v(\emptyset)}{d} \notag \\
        &= \frac{1}{d - 1} \left( \sum_{\emptyset \subsetneq \bm{S} \subsetneq \bm{N}, \atop i \in \bm{S}} d \omega(\bm{S}) v(\bm{S})
        \underbrace{\textcolor{blue}{- \sum_{\emptyset \subsetneq \bm{S} \subsetneq \bm{N}} \vert \bm{S} \vert \omega(\bm{S}) v(\bm{S})}}_{\text{Split into two terms}}  \right) + \frac{v(\bm{N}) - v(\emptyset)}{d} \notag \\
        &= \frac{1}{d - 1} \left( \sum_{\emptyset \subsetneq \bm{S} \subsetneq \bm{N}, \atop i \in \bm{S}} d \omega(\bm{S}) v(\bm{S})
        \textcolor{blue}{- \sum_{\emptyset \subsetneq \bm{S} \subsetneq \bm{N}, \atop i \in \bm{S}} \vert \bm{S} \vert \omega(\bm{S}) v(\bm{S}) - \sum_{\emptyset \subsetneq \bm{S} \subsetneq \bm{N}, \atop i \notin \bm{S}} \vert \bm{S} \vert \omega(\bm{S}) v(\bm{S})} \right) + \frac{v(\bm{N}) - v(\emptyset)}{d} \notag \\
        &= \frac{1}{d - 1} \left( \sum_{\emptyset \subsetneq \bm{S} \subsetneq \bm{N}, \atop i \in \bm{S}} (d - \vert \bm{S} \vert) \omega(\bm{S}) v(\bm{S})
        - \sum_{\emptyset \subsetneq \bm{S} \subsetneq \bm{N}, \atop i \notin \bm{S}} \vert \bm{S} \vert \omega(\bm{S}) v(\bm{S}) \right) + \frac{v(\bm{N}) - v(\emptyset)}{d} \notag \\
        &= \left( \sum_{\emptyset \subsetneq \bm{S} \subsetneq \bm{N}, \atop i \in \bm{S}} 
        \frac{(\vert \bm{S} \vert - 1)! (d - \vert \bm{S} \vert)!}{d!} v(\bm{S})
        - \sum_{\emptyset \subsetneq \bm{S} \subsetneq \bm{N}, \atop i \notin \bm{S}} 
        \frac{\vert \bm{S} \vert! (d - \vert \bm{S} \vert - 1)!}{d!} v(\bm{S}) \right) + \frac{v(\bm{N}) - v(\emptyset)}{d} \notag \\
        &= \sum_{\bm{S} \subset \bm{N}, \atop i \in \bm{S}} \frac{\vert \bm{S} \vert! (d - \vert \bm{S} \vert - 1)!}{d!} (v(\bm{S} \cup \{i\}) - v(\bm{S})). \tag{\text{Shapley values}} \label{equ:shap_value}
    \end{align}

    Above all, we can proof that the solution of \cref{equ:lsv} is equivalent to the Shapley values.

\end{proof}

\subsection{Proof of the unbiasness of \cref{equ:unbias_simshap}}

\Cref{equ:unbias_simshap} is unbiased in a probabilistic sense, as shown below:
\begin{align}
    \mathbb{E} [\phi_{\vx}] 
    &= \mT \sum_{\emptyset \subsetneq \bm{S} \subsetneq \bm{N}} p^{ls}(\bm{S}) a_{\bm{S}} v(\bm{S}) + \vb \notag \\
    &= \sum_{\emptyset \subsetneq \bm{S} \subsetneq \bm{N}} 
    \frac{[(d - |\bm{S}|)\mathbb{I}_{i\in \bm{S}} - |\bm{S}|\mathbb{I}_{i\notin \bm{S}}]v(\bm{S})}{\binom{d}{|\bm{S}|}(d - |\bm{S}|)|\bm{S}|} + \frac{v(\bm{N}) - v(\emptyset)}{d} \bm{1} \notag \\
    &= \sum_{\emptyset \subsetneq \bm{S} \subsetneq \bm{N}, i \in \bm{S}} 
    \frac{(d - |\bm{S}|) v(\bm{S})}{\binom{d}{|\bm{S}|}(d - |\bm{S}|)|\bm{S}|} 
    - \sum_{\emptyset \subsetneq \bm{S} \subsetneq \bm{N}, i \notin \bm{S}} 
    \frac{|\bm{S}| v(\bm{S})}{\binom{d}{|\bm{S}|}(d - |\bm{S}|)|\bm{S}|} 
    + \frac{v(\bm{N}) - v(\emptyset)}{d} \bm{1} \notag \\
    &= \sum_{\emptyset \subsetneq \bm{S} \subsetneq \bm{N}, i \in \bm{S}} 
    \frac{v(\bm{S})}{\binom{d}{|\bm{S}|}|\bm{S}|} 
    - \sum_{\emptyset \subsetneq \bm{S} \subsetneq \bm{N}, i \notin \bm{S}} 
    \frac{v(\bm{S})}{\binom{d}{|\bm{S}|}(d - |\bm{S}|)} 
    + \frac{v(\bm{N}) - v(\emptyset)}{d} \bm{1} \notag \\
    &= \sum_{\bm{S} \subset \bm{N} \setminus \{i\}} 
    \frac{\vert \bm{S} \vert! (d - \vert \bm{S} \vert - 1)!}{d!} (v(\bm{S} \cup \{i\}) - v(\bm{S})) \notag
\end{align}

\subsection{Proof of the equivalence of FastSHAP in \cref{equ:fastshap_equal}}

We simply expand the original equation as follows ($g$ represents $g(\vx;\theta)$ for short):
\begin{align}
    \mathcal{L} &= \mathop{\mathbb{E}}_{\vx \in \mathcal{X}}
    \left[ \left\Vert g - (\mU^T \mW \mU)^{-1} \mU^T \mW \vv_{\Delta} \right\Vert^2_{\mU^T \mW \mU} \right] \notag \\
    &= \mathop{\mathbb{E}}_{\vx \in \mathcal{X}} \left[
    (g - (\mU^T \mW \mU)^{-1} \mU^T \mW \vv_{\Delta})^T \mU^T \mW \mU (g - (\mU^T \mW \mU)^{-1} \mU^T \mW \vv_{\Delta}) \right] \notag \\
    &= \mathop{\mathbb{E}}_{\vx \in \mathcal{X}} \left[
        g^T \mU^T \mW \mU g - 2 g^T \mU^T \mW \vv_{\Delta} + \vv_{\Delta}^T \mW \mU (\mU^T \mW \mU)^{-1} \mU^T \mW \vv_{\Delta}
    \right] \notag \\
    &= \mathop{\mathbb{E}}_{\vx \in \mathcal{X}} \left[
        \Vert \mU g - \vv_{\Delta} \Vert_{\mW}^2 
    \right] + \underbrace{\mathop{\mathbb{E}}_{\vx \in \mathcal{X}} \left[
        - \vv_{\Delta}^T \mW \vv_{\Delta}
        + \vv_{\Delta}^T \mW \mU (\mU^T \mW \mU)^{-1} \mU^T \mW \vv_{\Delta}
    \right]}_{\text{Replace with constant $C$}} \notag \\
    &= \mathop{\mathbb{E}}_{\vx \in \mathcal{X}}
    \left[ \Vert \mU g - \vv_{\Delta} \Vert_{\mW}^2 \right] + C.
\end{align}
In summary, FastSHAP can be viewed as a special case of the unified amortized estimator proposed in this paper.

\subsection{Analysis of the potential noise of $\phi_x$ in \cref{equ:unified_amortied}}

We formulate this issue that for each input x, let the target be $\phi_x = \phi^{*} + \vn$, where the first term is the true Shapley value and the second is an input-agnostic noise term.
The noise term is independent of the input data (i.e., $p(\vx, \vn) = p(\vx) p(\vn)$) and the estimation is unbiased (i.e., $\mathbb{E}[\phi_x] = \phi^{*}$).
We can decompose the loss function in \cref{equ:unified_amortied} as follows:
\begin{align}
    L(\theta) &= \mathbb{E}\left[ \left\| g(\vx; \theta) - \phi_{\vx} \right\|_{\mM}^2 \right]\notag\\
    &=\mathbb{E}\left[ \left\| g(\vx; \theta) - \phi^{*} - \vn \right\|_{\mM}^2 \right]\notag\\
    &=\mathbb{E}\left[ \left\| g(\vx; \theta) - \phi^{*} \right\|_{\mM}^2 
        - 2 \left( g(\vx;\theta) - \phi^{*} \right)^T \mM \vn + \left\| \vn \right\|_{\mM}^2
    \right]\notag \\
    &=\mathbb{E}\left[ \left\| g(\vx; \theta) - \phi^{*} \right\|_{\mM}^2 \right]
    - \underbrace{\mathbb{E}\left[ 2 \left( g(\vx;\theta) - \phi^{*} \right)^T \right] \mM \mathbb{E} [\vn] }_{\text{Unbiased estimation $\mathbb{E}[\vn]=0$}}
    + \mathbb{E}\left[ \left\| \vn \right\|_{\mM}^2 \right]\notag \\
    &=\mathbb{E}\left[ \left\| g(\vx; \theta) - \phi^{*} \right\|_{\mM}^2 \right]
    + \mathbb{E}\left[ \left\| \vn \right\|_{\mM}^2 \right]\notag \\
    &=\mathbb{E}\left[ \left\| g(\vx; \theta) - \phi^{*} \right\|_{\mM}^2 \right] + C\notag\\
    &= \text{Error}(\theta) + C,
\end{align}
where $C = \mathbb{E}\left[ \left\|n \right\|_{\mM}^2 \right]$ and the metric matrix $\mM$ is positive definite.
In other words, minimizing $L(\theta)$ is equivalent to simply minimizing the Shapley value estimation error: mean-zero noise in the target doesn't affect the optimum, because it becomes a constant in the objective function.

\subsection{SimSHAP models and hyperparameters} \label{sec:details}

\paragraph{Tabular Datasets.}For the original model $f(x, \eta)$, we 
adopted LIGHTGBM~\citep{ke2017lightgbm} for implementation. The surrogate model
is implemented using 3 to 6-layer MLPs with 128/512 hidden units and ELU activations.
 The SimSHAP explainer model is implemented using neural networks that consist of 3 fully connected layers with 128/512 units and ReLU activations, 
 according to the input feature of the dataset. It is trained AdamW optimizer with learning rate of range $7\times 10^{-4}$ to $1.5\times 10^{-2}$ and batch size of 1024/2048. All the models don't contain Softmax Layers. All the baselines for SimSHAP accuracy comparison were computed using an open-source implementation
\footnote{\href{https://github.com/iancovert/shapley-regression/}{https://github.com/iancovert/shapley-regression/} (License: MIT)}. 
and FastSHAP 
package\footnote{\href{https://github.com/iancovert/fastshap/}
{https://github.com/iancovert/fastshap}(License: MIT)}. All the experiments were run on a GEFORCE RTX 3090 Card.

\paragraph{Image Datasets.}The original model and surrogate model are both ResNet-18 structures. 
Specifically, the surrogate model takes 2 inputs, i.e., an image and a mask, and returns an output of 10 dimensions.
The SimSHAP explainer model is implemented using a U-Net strucuture, with the output of $Batch Size \times y \times 16\times 16$.
It is trained using AdamW optimizer with a learning rate of $2\times 10^{-4}$ and batch size of 256.
All the models don't contain Softmax Layers.
To ensure comparability of all the methods, we average-pooled the results of methods that provide attributions of the same size as the image. Integrated-Gradients and SmoothGrad
 are implemented through \texttt{captum} package\footnote{\href{https://captum.ai/docs/introduction}{https://captum.ai/docs/introduction/ }(License: MIT)}
, KernelSHAP, KernelSHAP-S, and DeepSHAP are implemented through \texttt{shap} package\footnote{\href{https://shap.readthedocs.io/en/latest/}{https://shap.readthedocs.io/en/latest/ }(License: MIT)}, and GradCAM
  is implemented through an open-source package for explanation benchmarks\footnote{\href{https://github.com/zbr17/ExplainAttr}{https://github.com/zbr17/ExplainAttr/ }(License: MIT)}.

\begin{figure}[htbp]
    \centering
    
    \begin{subfigure}{0.33\textwidth}
        \includegraphics[width=1\textwidth]{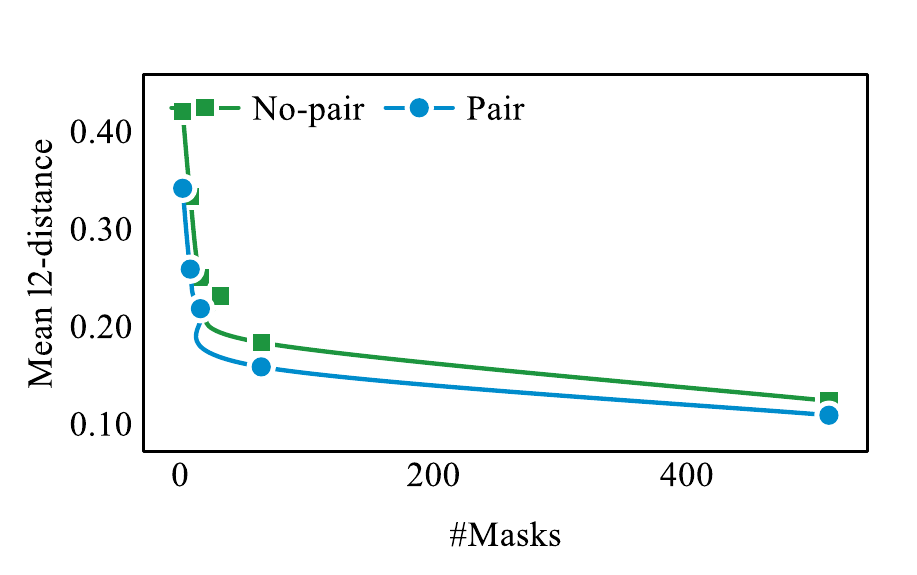}
        \caption{}
        \label{fig:census_mask}
    \end{subfigure}
    \begin{subfigure}{0.33\textwidth}
        \includegraphics[width=1\textwidth]{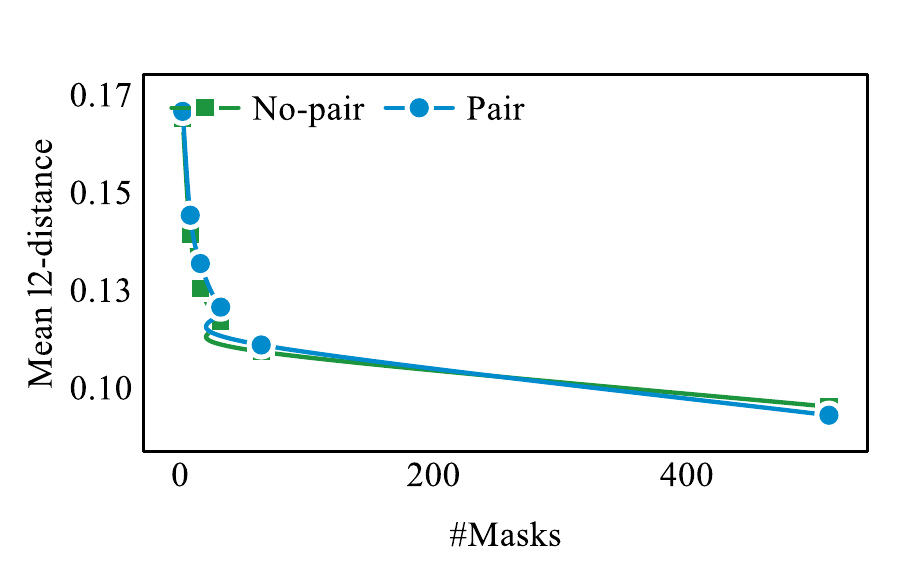}
        \caption{}
        \label{fig:news_mask}
    \end{subfigure}
    \begin{subfigure}{0.32\textwidth}
        \includegraphics[width=1\textwidth]{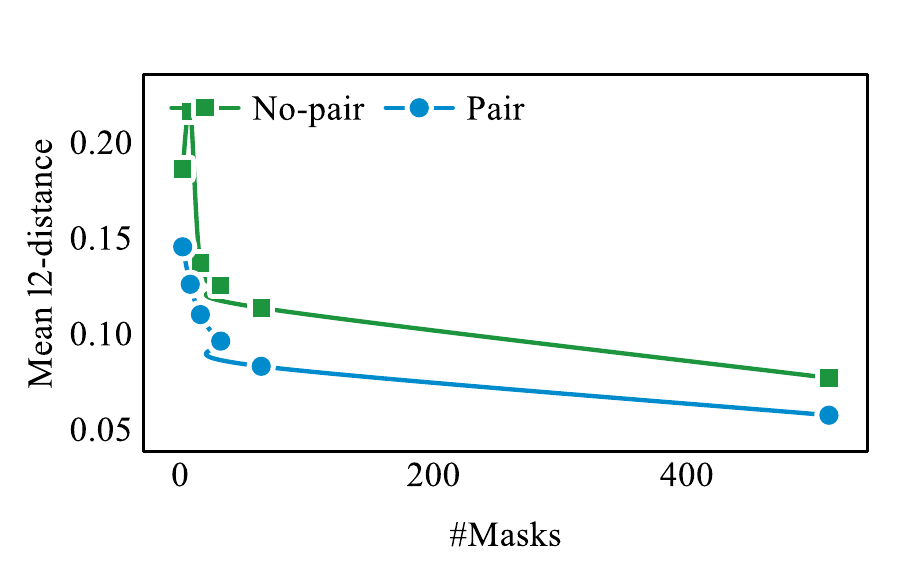}
        \caption{}
        \label{fig:bank_mask}
    \end{subfigure}
    \caption{SimSHAP accuracy as a function of number of training samples with/out pair sampling in (a) Census (b) News (c) Bank dataset.}
    \label{fig:mask}
\end{figure}

\cref{fig:mask} shows SimSHAP accuracy as a function of the number of training samples. Results reveal that across all tabular datasets,
increasing the number of masks improves accuracy. However, as the number of masks increases, the accuracy gain becomes less significant.
Additionally, pair sampling at least doesn't make the accuracy worse. In fact, pair sampling can improve accuracy in most cases.

\subsection{Ablation Studies}
\label{sec:ablation}
\paragraph{Tabular Dataset}
\begin{table}
    \centering
    \begin{minipage}{0.48\textwidth}\centering
            \small
            \tabcolsep=0.1cm
            \caption{Influence of learning rate.}

            \begin{tabular}{ccc}
        \toprule
            & \textbf{Mean $l_1$ distance} & \textbf{Mean $l_2$ distance} \\
    \cmidrule{2-3}    $1\times 10^{-1}$ & 1.587  & 0.197  \\
        $1\times 10^{-2}$ & 2.051  & 0.224  \\
        $1\times 10^{-3}$ & 0.926  & 0.106  \\
        $7\times 10^{-4}$ & \textbf{0.886}  & \textbf{0.098}  \\
        $1\times 10^{-5}$ & 2.084  & 0.194  \\
        $1\times 10^{-6}$ & 5.950  & 0.548  \\
        \bottomrule
        \label{tab:ablation_bank_lr}
        \end{tabular}%
    \end{minipage}
    \begin{minipage}{0.48\textwidth}\centering
        \small
        \tabcolsep=0.1cm
        \vspace{-8.5mm}
        \caption{Influence of optimization epoch.}

                \begin{tabular}{ccc}
        \toprule
            & \textbf{Mean $l_1$ distance} & \textbf{Mean $l_2$ distance} \\
    \cmidrule{2-3}    100   & 1.550  & 0.152  \\
        500   & 1.009  & 0.107  \\
        1000  & 0.886  & 0.098  \\
        1500  & \textbf{0.824}  & \textbf{0.094}  \\
        \bottomrule
        \label{tab:ablation_bank_epoch}
        \end{tabular}%

    \end{minipage}
\end{table}

\begin{table}
    \begin{minipage}{0.48\textwidth}\centering
            \small
            \tabcolsep=0.1cm
            \caption{Influence of batch size.}
        \begin{tabular}{ccc}
        \toprule
            & \textbf{Mean $l_1$ distance} & \textbf{Mean $l_2$ distance} \\
    \cmidrule{2-3}    32    & 1.430  & 0.164  \\
        128   & 1.176  & 0.129  \\
        512   & 0.933  & 0.103  \\
        1024  & 0.886  & 0.098  \\
        Whole Set & \textbf{0.857}  & \textbf{0.092}  \\
        \bottomrule
        \label{tab:ablation_bank_batchsize}
        \end{tabular}%

    \end{minipage}
    \begin{minipage}{0.48\textwidth}\centering
        \small
        \tabcolsep=0.1cm
        \vspace{-10.5mm}
        \caption{Influence of net structure}

    \begin{tabular}{lcc}
        \toprule
            & \multicolumn{1}{c}{\textbf{Mean $l_1$ distance}} & \multicolumn{1}{c}{\textbf{Mean $l_2$ distance}} \\
    \cmidrule{2-3}    wider & 0.933  & 0.108  \\
        deeper & \textbf{0.865}  & \textbf{0.099}  \\
        \bottomrule
        \label{tab:ablation_bank_net}
        \end{tabular}%

    \end{minipage}
\end{table}

We delve into an ablation study focusing on hyperparameters on \texttt{bankruptcy} dataset.
We use the default configuration of 32 samples, pair sampling, and 1000 epochs for this study. \cref{tab:ablation_bank_lr,tab:ablation_bank_epoch,tab:ablation_bank_batchsize,tab:ablation_bank_net} presents the results. For learning rate, we observed that the best performance is achieved with learning rates ranging from
$1\times 10^{-3}$ and $1\times 10^{-5}$ for 1000-epoch itereations. 
A high learning rate can lead to an unstable training process, while a low learning rate can lead to slow convergence. Therefore, in order to 
achieve the $O(\frac{1}{\sqrt{M}})$ convergence, the learning rate should be chosen carefully.
As for batch size, we find that larger batch sizes result in improved performance.
As a result, the batch size can be increased as long as it remains within the constraints of available GPU memory.
We evaluated performance across 100, 500, 1000, 1500 iterations. While performance improves with higher iteration counts, gains become less significant after 500 epochs. 
Striking a balance between accuracy and training time, we chose 1000 iterations for this dataset.
We also explored the architecture of the explainer model. For MLPs, increasing the number of layers for a deeper model and increasing the hidden dimension for a wider model were considered. Results indicate that a 3-layer MLP with a hidden dimension of 512 is sufficient in capturing the Shapley space of raw data.

\paragraph{Image Dataset}
\begin{table}
\begin{minipage}{0.45\textwidth}\centering
        \small
        \caption{Mean Insertion AUC and Deletion AUC of SimSHAP as a function of learning rate.}
    \begin{tabular}{ccc}
    \toprule
          & \textbf{Ins. AUC} & \textbf{Del. AUC} \\
\cmidrule{2-3}    $1\times 10^{-2}$ & 0.667  & \textbf{-0.365} \\
    $1\times 10^{-3}$ & 0.733  & -0.301 \\
    $2\times 10^{-4}$ & \textbf{0.755 } & -0.302 \\
    $1\times 10^{-5}$ & 0.691  & -0.302 \\
    $1\times 10^{-6}$ & 0.498  & -0.188 \\
    \bottomrule
    \end{tabular}%
  \label{tab:ablation_cifar_lr}%

\end{minipage}
\hspace{0.1\textwidth}
\begin{minipage}{0.45\textwidth}\centering
    \small
    \caption{Mean Insertion AUC and Deletion AUC of SimSHAP as a function of epoch.}
    \begin{tabular}{ccc}
    \toprule
          & \textbf{Ins. AUC} & \textbf{Del. AUC} \\
\cmidrule{2-3}    200   & 0.726 & -0.332 \\
    400   & 0.740  & -0.312 \\
    600   & \textbf{0.777} & -0.288 \\
    800   & 0.720  & -0.302 \\
    1000  & 0.703 & \textbf{-0.348} \\
    \bottomrule
    \end{tabular}%
  \label{tab:ablation_cifar_epoch}%

\end{minipage}
\end{table}
\begin{table}
\begin{minipage}{0.45\textwidth}\centering
        \small
        \caption{Mean Insertion AUC and Deletion AUC of SimSHAP as a function of batch size.}
    \begin{tabular}{ccc}
    \toprule
          & \textbf{Ins. AUC} & \textbf{Del. AUC} \\
\cmidrule{2-3}    8     & 0.721 & -0.335 \\
    16    & 0.685 & -0.277 \\
    32    & 0.703 & \textbf{-0.337} \\
    64    & 0.716 & -0.298 \\
    256   & \textbf{0.755} & -0.302 \\
    \bottomrule
    \end{tabular}%
  \label{tab:ablation_cifar_batchsize}%

\end{minipage}
\hspace{0.1\textwidth}
\begin{minipage}{0.45\textwidth}\centering
    \small
    \vspace{6mm}
    \caption{Mean Insertion AUC and Deletion AUC of SimSHAP as a function of limited data.}
    \begin{tabular}{ccc}
    \toprule
    \multicolumn{1}{l}{\textbf{Percent(\%)}} & \multicolumn{1}{l}{\textbf{Ins. AUC}} & \multicolumn{1}{l}{\textbf{Del. AUC}} \\
    \midrule
    4     & 0.532  & -0.168  \\
    16    & 0.736  & \textbf{-0.359}  \\
    24    & 0.695  & -0.357  \\
    40    & \textbf{0.767}  & -0.298  \\
    48    & 0.749  & -0.325  \\
    64    & 0.760  & -0.337  \\
    80    & 0.702  & -0.323  \\
    100   & 0.755  & -0.302 \\
    \bottomrule
    \end{tabular}%
  \label{tab:ablation_cifar_limiteddata}\end{minipage}
\end{table}
Similarly, we explored the hyperparameters of the \textbf{CIFAR-10} dataset.
We use the default configuration of 8 samples, pair sampling, learning rate of $2\times 10^{-4}$, and 500 epochs for this study.
We first utilize \textbf{CIFAR-10} dataset to evaluate the performance of SimSHAP via the size of the dataset.
Results in \cref{tab:ablation_cifar_limiteddata} show the robustness of SimSHAP for achieving great performance with a mere 20\% of training data. 
When more data is involved, SimSHAP can have small improvements but contribute to lower variance regarding the AUC.

Detailed comparison are in \cref{tab:ablation_cifar_lr,tab:ablation_cifar_batchsize,tab:ablation_cifar_epoch}. For learning rate, we observe that the best performance is achieved with learning rates around $1\times 10^{-4}$. 
For the number of epochs, we find that 500-600 epochs is sufficient to balance the training speed and accuracy. More epochs may 
lower the variance for Inclusion and Deletion scores, but the improvement is not significant.
For the number of samples, we only tested the pair sampling cases. Results reveal that 
when the number of samples exceeds 8, the improvement is not significant, and there is a slight decrease in performance. This laid 
a foundation for a relatively fast and accurate training method with a small number of samples on image datasets.
For batch size, similar to tabular datasets, a larger batch size may lead to better performance. However the parameter also 
needs to be carefully chosen for the GPU memory constraints.


\subsection{Training on Inexact Labels on Iris Dataset}
\label{sec:iris_dataset_appendix}
In this section, we demonstrate SimSHAP's ability to train on inexact labels by conducting experiments on the Iris dataset. It's crucial to note that SimSHAP bypasses the need of ground truth labels for training by estimating them using limited sample data, which is also true for FastSHAP's Least Squares value~\citep{jethani2021fastshap}. The size of the dataset should be sufficiently large, or else the model might overfit to the noise introduced during the estimation process.

\cref{tab:iris} demonstrates the results of the mean $l_2$ distance between the model's output and the estimated ground truth label, as the dataset size increases. Each configuration was trained until convergence. Together with the result in \cref{tab:ablation_cifar_limiteddata}, we can conclude that SimSHAP is indeed capable of learning from noisy data when provided with a sufficiently large dataset. When the dataset is small, there is a higher likelihood that the model will learn the noise rather than the true Shapley values.
\begin{table}[htbp]
  \centering
  \caption{Mean $l_2$ distance as a function of data size}
\begin{tabular}{cc}
\toprule
\multicolumn{1}{l}{\textbf{Size of Dataset}} & \textbf{Distance} \\
\midrule
5     & 0.577 \\
20    & 0.144 \\
45    & 0.070 \\
60    & 0.052 \\
75    & 0.050 \\
100   & 0.042 \\
120 (whole) & 0.033 \\
\bottomrule
\end{tabular}%
\label{tab:iris}%
\end{table}%

\subsection{An Exact Estimator for Unbiased KernelSHAP}
\label{sec:unbiased}

Following Section 3.2 of the unbiased KernelSHAP estimation~\citep{covert2021improving}, we provide ``an approximate solution to the exact problem'' with the Lagrangian as follows:
\begin{align}
L(\eta, \lambda) &= \eta^T \mathbb{E}\left[(\bm{1}^{\bm{S}})(\bm{1}^{\bm{S}})^T\right]\eta \notag\\
&-2\eta^T\mathbb{E}\left[\bm{1}^{\bm{S}}(v(\bm{S}) - v(\emptyset))\right] \notag\\
&+\mathbb{E}\left[(v(\bm{S}) - v(\emptyset))^2\right]\notag\\
&+\lambda(\bm{1}^T\eta- v(\bm{N}) + v(\emptyset)).
\end{align}
Using the shorthand notation
\begin{align}
    A = \mathbb{E}\left[(\bm{1}^{\bm{S}})(\bm{1}^{\bm{S}})^T\right], ~~b = \mathbb{E}\left[\bm{1}^{\bm{S}}(v(\bm{S}) - v(\emptyset))\right],\notag
\end{align}
we can calculate $A$ precisely and only need to estimate $b$ by Monte Carlo Sampling:
\begin{align}
    \bar{b}_M = \frac{1}{M}\sum_{k=1}^M \bm{1}^{\bm{S_k}}v(\bm{S_k}) - \mathbb{E}\left[\bm{1}^{\bm{S}}\right]v(\emptyset).
\end{align}
The unbiased KernelSHAP is formulated as follows:
\begin{align}
    \eta_M = A^{-1}(\bar{b}_M - \bm{1}\frac{\bm{1}^TA^{-1}\bar{b}_M - v(\bm{N})+v(\emptyset)}{\bm{1}^TA^{-1}\bm{1}}).\
\end{align}
We recommend readers to refer to Section 1 of the Supplementary material in~\citet{covert2021improving}.

\subsection{Detailed Comparison Between Our Work and ~\citet{schwarzenberg2021efficient,chuang2023cortx}}
\label{sec:literature_comparison}
\paragraph{About the Similarity}
We acknowledge the similarities between our work and ~\citet{schwarzenberg2021efficient,chuang2023cortx}, but it should be noted that both works not only deal with Shapley values computation. Firstly, ~\citet{schwarzenberg2021efficient} proposed a general framework based on the concept of amortized estimation, in which a single neural network is trained to predict the Integrated Gradients / Shapley values for each player (either a pixel in an image or a token in text). Secondly, ~\citet{chuang2023cortx} is rooted in RTX, which is fundamentally equivalent to amortized estimation (with FastSHAP mentioned as the first line of work or the RTX paradigm).
\paragraph{About the Difference}
However, there are also notable distinctions. Schwarzenberg et al ensures accuracy in the first term of their Eq.1: $argmin_{\theta\in \Theta}\frac{1}{\left|X\right|}\sum_{x\in \textbf{X}}\alpha D(E_f(x), e_{\theta}(x)) + \beta(\frac{\left\|e_{\theta}(x)\right\|}{\left\|E_{f}(x)\right\|})$. If we were to utilize ground truth labels, our framework would align with this approach, matching the $E_f(x)$ of the above equation, indicating the expensive explainer. Nevertheless, we didn't apply any form of supervision, instead relying solely on sampled ground truth data for our MSE computations. For ~\citet{chuang2023cortx}, it is mentioned in the Introduction section of CoRTX that methods like Fastshap "learn an explainer to minimize the estimation error regarding to the approximated explanation labels". We concur with this assessment, though this work does not provide a definitive explanation on this matter. FastSHAP, based on \cref{equ:fastshap_equal}, represents a specific example of approximation with a complicated metric matrix, and these 2 works can also be seen as specific examples of our proposed framework. Furthermore, both the fine-tuning stage of CoRTX and the Supervised RTX baseline they used require ground truth labels for training, which is not the case in SimSHAP. Similarly to FastSHAP, we do not acquire any ground truth labels during training.

\subsection{Analysis of the Time Complexity of FastSHAP and SimSHAP Algorithms}
\label{sec:algorithm_complexity}
In addition to the unified framework, we also want to highlight that the time complexity of SimSHAP and FastSHAP is comparable, which corresponds to the amortized estimation version of the semi-value and least squares value.

We define $B$ as the number of batches, $d$ as the input feature dimension, $K$ as the number of samples, and $o$ as the output dimension of the classifier (which refers to a number of classes).

We can rewrite the main loss function of FastSHAP as follows:
\begin{align}
    L_{fastshap} = \sum_{b=1}^{B}\sum_{y=1}^o\sum_{k=1}^K(v_y(\bm{S}_k)-v_y(\emptyset) - \bm{S}_k^T\hat{\phi})^2,\notag
\end{align}
where $v_y(\cdot)$ indicates the yth  component of vector $v(\cdot)$. Ignoring the computation cost of $v(\bm{S}_k)-v(\emptyset)$, the number of multiplications is $(d+2)KoB$, and the number of additions is $dKoB$.

Similarly, we can rewrite the main loss function of SimSHAP as follows:
\begin{align}
    L_{simshap} = \sum_{b=1}^{B}(\hat{\phi} - \phi_{sample})^2,
\end{align}
where 
\begin{align}
    \phi_{sample} = \sum_{k=1}^K \omega({\bm{S}_k})v^T(\bm{S}_k) + \frac{v(\bm{N}) - v(\emptyset)}{d}.
\end{align}
Note that $\hat{\phi}$ has the dimension of $o\times d$. Ignoring the computation cost of $\frac{v(\bm{N}) - v(\emptyset)}{d}$, the number of multiplications is $(dKo+1)B$, the number of additions is $(2do+K-1)B$.

According to the above comparison, the time complexities of the two algorithms are $\mathcal{O}(dKoB)$ under the framework of Big O notation.



\end{document}